\documentclass[10pt,journal,compsoc]{IEEEtran}
\ifCLASSOPTIONcompsoc
  \usepackage[nocompress]{cite}
\else
  \usepackage{cite}
\fi
\ifCLASSINFOpdf
  \usepackage[pdftex]{graphicx}
  \graphicspath{{../figures/}}
  \DeclareGraphicsExtensions{.pdf}
\else
\fi
\usepackage{amsmath}
\usepackage{amssymb}
\usepackage{url}

\usepackage{booktabs}

\begin{document}
%
\title{Smoothed Dilated Convolutions for Improved Dense Prediction}
%
%
%
%

\author{Zhengyang~Wang
        and~Shuiwang~Ji,~\IEEEmembership{Senior~Member,~IEEE}
\IEEEcompsocitemizethanks{\IEEEcompsocthanksitem Zhengyang Wang and Shuiwang Ji are with the Department of Computer Science and Engineering, Texas A\&M University, College Station, TX, 77843.\protect\\
E-mail: sji@tamu.edu}
\thanks{Manuscript received April 30, 2019.}}

%
%

\markboth{IEEE Transactions on Pattern Analysis and Machine Intelligence,~Vol.~xx, No.~x, April~2019}%
{Wang \MakeLowercase{\textit{et al.}}: Smoothed Dilated Convolutions for Improved Dense Prediction}
%



\IEEEtitleabstractindextext{%
\begin{abstract}
Dilated convolutions, also known as atrous convolutions, have been widely explored in deep convolutional neural networks~(DCNNs) for various dense prediction tasks. However, dilated convolutions suffer from the gridding artifacts, which hampers the performance. In this work, we propose two simple yet effective degridding methods by studying a decomposition of dilated convolutions. Unlike existing models, which explore solutions by focusing on a block of cascaded dilated convolutional layers, our methods address the gridding artifacts by smoothing the dilated convolution itself. In addition, we point out that the two degridding approaches are intrinsically related and define separable and shared~(SS) operations, which generalize the proposed methods. We further explore SS operations in view of operations on graphs and propose the SS output layer, which is able to smooth the entire DCNNs by only replacing the output layer. We evaluate our degridding methods and the SS output layer thoroughly, and visualize the smoothing effect through effective receptive field analysis. Results show that our methods degridding yield consistent improvements on the performance of dense prediction tasks, while adding negligible amounts of extra training parameters. And the SS output layer improves the performance significantly and is very efficient in terms of number of training parameters.
\end{abstract}

}

\maketitle

\IEEEdisplaynontitleabstractindextext

%
\IEEEpeerreviewmaketitle

\IEEEraisesectionheading{\section{Introduction}\label{sec:intro}}

\begin{figure}[!t]
\centering
\includegraphics[width=0.48\textwidth]{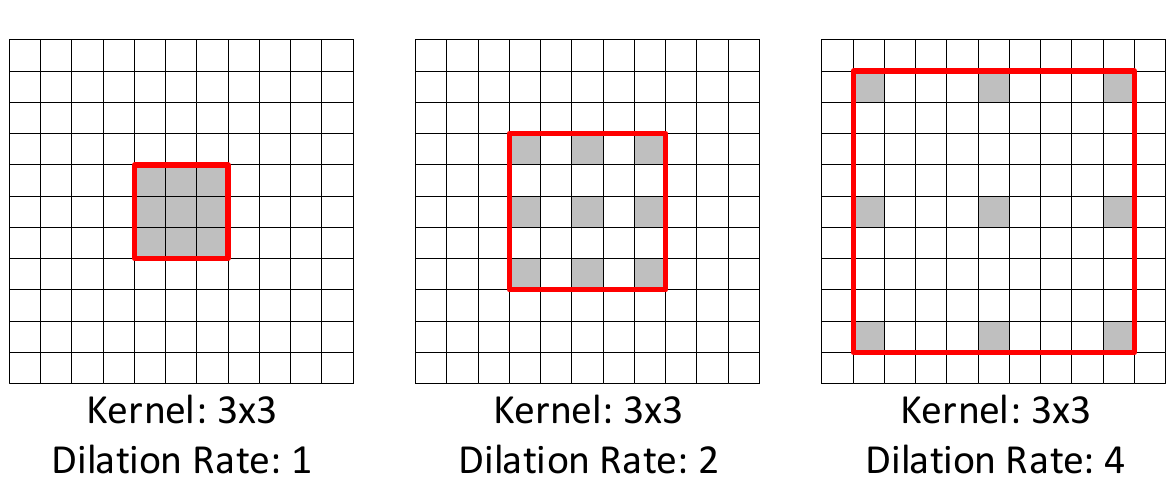}
\caption{$2$-D Dilated convolutions with a kernel size of $3 \times
	3$. Note that when the dilation rate is $1$, dilated convolutions
	are the same as standard convolutions. Dilated convolutions enlarge
	the receptive field while keeping the spatial resolution.}
\label{fig:dilatedconv}
\end{figure}

\IEEEPARstart{D}{ilated} convolutions, also known as atrous convolutions, have been
widely explored in deep convolutional neural networks~(DCNNs) for
various tasks, including semantic image
segmentation~\cite{giusti2013fast,li2014highly,yu2015multi,yu2017dilated,chen2016deeplab,chen2017rethinking,wang2017understanding,hamaguchi2017effective,zhao2016pyramid,gao2017pixel},
object
detection~\cite{sermanet2013overfeat,papandreou2015modeling,dai2016r,huang2016speed},
audio generation~\cite{oord2016wavenet}, video
modeling~\cite{kalchbrenner2016video}, and machine
translation~\cite{kalchbrenner2016neural}. The idea of dilated
filters was developed in the \textit{algorithm \`{a} trous} for
efficient wavelet decomposition in~\cite{holschneider1990real} and
has been used in image pixel-wise prediction tasks to allow
efficient
computation~\cite{giusti2013fast,li2014highly,sermanet2013overfeat,papandreou2015modeling}.
Dilation upsamples convolutional filters by inserting zeros between
weights, as illustrated in Figure~\ref{fig:dilatedconv}. It enlarges
the receptive field, or field of
view~\cite{chen2016deeplab,chen2017rethinking,hamaguchi2017effective},
but does not require training extra parameters in DCNNs. Dilated
convolutions can be used in cascade to build multi-layer networks
\cite{oord2016wavenet,kalchbrenner2016video,kalchbrenner2016neural}.
Another advantage of dilated convolutions is that they do not reduce
the spatial resolution of responses. This is a key difference from
downsampling layers, such as pooling layers or convolutions with
stride larger than one, which also expand the receptive field of
subsequent layers but also reduce the spatial resolution. This
allows the transfer of classification models trained on
ImageNet~\cite{deng2009imagenet,he2016deep} to semantic image
segmentation tasks by removing downsampling layers and applying
dilation in convolutions of subsequent
layers~\cite{long2015fully,yu2015multi,yu2017dilated,chen2016deeplab,chen2017rethinking,wang2017understanding,hamaguchi2017effective,zhao2016pyramid}.
Similar to standard convolutions, a layer consisting of a dilated
convolution with an activation function is called a dilated
convolutional layer.

While DCNNs with dilated convolutions achieved success in a wide
variety of deep learning tasks, it has been observed that dilations
result in the so-called ``gridding
artifacts''~\cite{yu2017dilated,wang2017understanding,hamaguchi2017effective}.
For dilated convolutions with dilation rates larger than one,
adjacent units in the output are computed from completely separate
sets of units in the input. It results in inconsistency of local
information and hampers the performance of DCNNs with dilated
convolutions. As dilated convolutional layers are commonly stacked
together in cascade in DCNNs, existing models focus on smoothing such
gridding artifacts for a block of cascaded dilated convolutional layers.
In~\cite{yu2017dilated,hamaguchi2017effective} the gridding problem was
alleviated by adding more layers with millions of extra training
parameters after the block of dilated convolutions.
In~\cite{wang2017understanding} the hybrid dilated convolution~(HDC)
was proposed, which applies different dilation rates without a common
factor for continuous dilated convolutional layers.

In this work, we address the gridding artifacts by smoothing the
dilated convolution itself, instead of a block of stacked dilated convolutional layers. Our
methods enjoy the unique advantage of being able to replace any
single dilated convolutional layer in existing networks as they do not rely on
other layers to solve the gridding problem. More importantly, our
methods add minimal numbers of extra parameters to the model while
some other degridding approaches increase the model parameters
dramatically~\cite{yu2017dilated,hamaguchi2017effective}.
Our methods are based on an interesting view of the dilated
convolutional
operation~\cite{shensa1992discrete,chen2016deeplab,abadi2016tensorflow},
which benefits from a decomposition of the operations. Based on this
novel interpretation of dilated convolutions, we propose two simple
yet effective methods to smooth the gridding artifacts. By analyzing
these two methods in both the original operation and
the decomposition views, we further notice that they are
intrinsically related and define separable and shared~(SS)
operations that generalize the proposed methods. Experimental
results show that our methods improve current DCNNs with dilated
convolutions significantly and consistently, while only adding a few
hundred extra parameters. We also employ the effective receptive
field~(ERF) analysis~\cite{luo2016understanding} to visualize the
smoothing effect for DCNNs with our dilated convolutions.

Compared to the original conference version of this
paper~\cite{wang2018smoothed}, we perform further analysis on SS
operations in view of operations on graphs. Based on this analysis,
we incorporate deep learning techniques on graphs and propose the SS
output layer, which smooths DCNNs with dilated convolutions by only
replacing the output layer. In addition, the SS output layer shows a
better ability of aggregating information from large receptive
fields than original output layers based on dilated convolution. The
smoothed DCNNs are able to produce significantly improved dense
prediction.

\section{Background and Related Work}\label{sec:bg}

In this section, we describe dilated convolutions and DCNNs with them. We
then discuss the gridding problem and current solutions in detail.

\subsection{Dilated Convolutions}\label{sec:dilatedconv}

In the one-dimensional case, given a $1$-D input~$f$, the output~$o$
at location $i$ of a dilated convolution with a filter~$w$ of
size~$S$ is defined as
\begin{equation}\label{eqn:1}
o[i]=\sum_{s=1}^{S}f[i+r \cdot s]w[i],
\end{equation}
where $r$ is known as the dilation rate. Higher dimensional cases
can be easily generalized. When $r=1$, dilated convolutions
correspond to standard convolutions. An intuitive and direct way to
understand dilated convolutions is that $r-1$ zeros are inserted
between every two adjacent weights in the standard convolutional
filters. Dilated convolutions are also known as atrous convolutions
in which ``trous'' means holes in French.
Figure~\ref{fig:dilatedconv} contains an illustration of dilated
convolution in the two-dimensional case.

As mentioned in Section~\ref{sec:intro}, in most cases, DCNNs use
dilated convolutions in cascade, which means several dilated
convolutional layers are stacked together. The reasons for using
this cascaded pattern differ for different tasks. In the task of
semantic image
segmentation~\cite{long2015fully,yu2015multi,yu2017dilated,chen2016deeplab,chen2017rethinking,wang2017understanding,hamaguchi2017effective,zhao2016pyramid},
in order to have output feature maps of larger sizes while
maintaining the size of the receptive field, dilated convolutions
are employed to replace standard convolutions in layers after the
removed downsampling layers. For example, if we treat standard
convolutions as dilated convolutions with a dilation rate of $r=1$,
when a downsampling layer with a subsampling rate of $2$ is removed,
the dilation rates of all subsequent convolutional layers should be
multiplied by $2$. This results in dilated convolutional layers with
dilation rates of $r=2, 4, 8,$ etc. In other tasks, such as audio
generation~\cite{oord2016wavenet}, video
modeling~\cite{kalchbrenner2016video}, and machine
translation~\cite{kalchbrenner2016neural}, the use of dilated
convolutions aims at enlarging the receptive fields of outputs. As
pointed out
in~\cite{yu2015multi,oord2016wavenet,kalchbrenner2016video},
cascaded dilated convolutional layers expand the receptive field
exponentially in the number of layers in DCNNs, as opposed to
linearly. In these studies, the dilation rate is doubled for every
forward layer, starting from $1$ up to a limit before the pattern is
repeated.

Note that when using dilated convolutions in cascade, the gridding
artifacts affect the models more significantly. This is because the
dilation rates of continuously stacked layers have a common factor
of $2$ in all of these DCNNs that use dilated convolutional layers
in cascade, as discussed in~\cite{wang2017understanding} and
Section~\ref{sec:gridding}. In~\cite{chen2016deeplab,chen2017rethinking}
dilated convolutions in parallel to form the output layer
were explored.

\subsection{Gridding in Dilated Convolutions}\label{sec:gridding}

\begin{figure}[!t]
\centering
\includegraphics[width=0.48\textwidth]{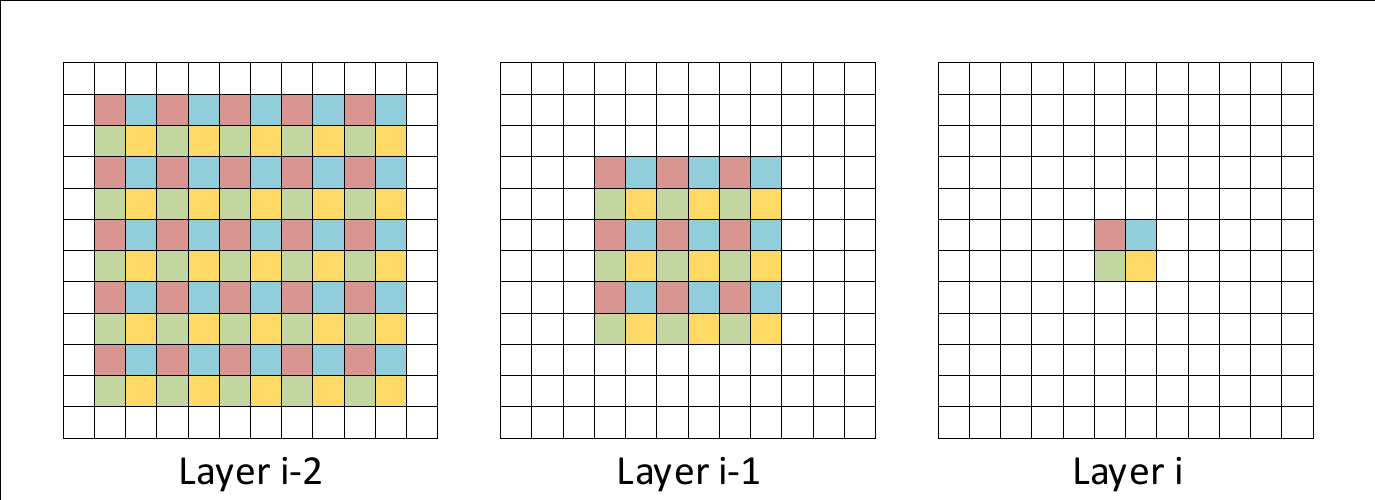}
\caption{An illustration of gridding artifacts. The operations
	between layers are both dilated convolutions with a kernel size of
	$3 \times 3$ and a dilation rate of $r=2$. For four neighboring
	units in layer $i$ indicated by different colors, we mark their
	actual receptive fields in layer $i-1$ and $i-2$ using the same
	color, respectively. Clearly, their actual receptive fields are
	completely separate sets of units.}
\label{fig:gridding}
\end{figure}

Dilated convolutions with dilation rates larger than one will
produce the so called gridding artifacts; that is, adjacent units in
the output are computed from completely separate sets of units in
the input and thus have totally different actual receptive fields.
To view the gridding problem clearly, we first look into a single
dilated convolution. Considering the second case in
Figure~\ref{fig:dilatedconv} as an example, a $2$-D dilated
convolution with a kernel size of $3 \times 3$ and a dilation rate
of $r=2$ has a $5 \times 5$ receptive field. However, the number of
pixels that are actually involved in the computation is only $9$ out
of $25$, which implies that the actual receptive field is still $3
\times 3$, but sparsely distributed. If we further consider the
neighboring units in the output, the gridding problem can be seen
from Figure~\ref{fig:gridding}. Suppose we have two consecutive
dilated convolutional layers in cascade, and both dilated
convolutions have a kernel size of $3 \times 3$ and a dilation rate
of $r=2$. For four adjacent units indicated by different colors in
layer $i$, we show their actual receptive fields in layer $i-1$ and
$i-2$ using the same color. We can see that four completely separate
sets of units in layer $i-1$ contribute to the computation of the
four units in layer $i$. Moreover, since the dilation rates for both
layers are $2$, which have a common factor of $2$, the gridding
problem also exists in layer $i-2$. Indeed, whenever the dilation
rates of dilated convolutional layers in cascade have a common
factor relationship, such as $2,2,2$ or $2,4,8$, the gridding
problem is propagated to all layers, as pointed out
in~\cite{wang2017understanding}. For a block of such layers,
neighboring outputs of the block are computed from totally different
sets of inputs. This results in the inconsistency of local
information and hampers the performance of DCNNs with dilated
convolutions.

The gridding artifacts were observed and addressed in several recent
studies for semantic image
segmentation~\cite{yu2017dilated,wang2017understanding,hamaguchi2017effective}.
As described in Section~\ref{sec:dilatedconv}, dilated convolutions
are mostly employed in cascade in DCNNs. Therefore, these studies
focused on solving the gridding problem in terms of a block of stacked
dilated convolutional layers. Specifically, hybrid dilated
convolution~(HDC) was proposed in \cite{wang2017understanding},
which groups several dilated convolutional layers and applies
dilation rates without a common factor relationship. For example,
for a block of dilated convolutions with a dilation rate of $r=2$,
every three consecutive layers are grouped together and the
corresponding dilation rates are changed to $1,2,3$ instead of
$2,2,2$. For a similar block with a dilation rate of $r=4$, the same
grouping principle is applied and the dilation rates become $3,4,5$,
instead of $4,4,4$. When used together with their proposed dense
upsampling convolution~(DUC), this approach improved DCNNs for
semantic image segmentation. This strategy was also adopted as the
``multigrid'' method in recent work~\cite{chen2017rethinking}. Prior
to~\cite{wang2017understanding}, the degridding was performed mainly
by adding more layers after the block of dilated convolutional
layers~\cite{yu2017dilated,hamaguchi2017effective}. It was proposed
in \cite{yu2017dilated} to add two more standard convolutional
layers without residual connections while
\cite{hamaguchi2017effective} proposed to add a block of dilated
convolutional layers with decreasing dilation rates. The main
drawback of such methods is the requirement for learning a large
amount of extra parameters.

\section{Smoothed Dilated Convolutions}\label{sec:main}

In this section, we discuss a decomposition view of dilated
convolutions. We then propose two approaches for smoothing the
gridding artifacts. We also analyze the relationship between the
proposed two methods and define separable and shared~(SS) operations
to generalize them. Based on this analysis, we
further propose the SS output layer to perform degridding for the
entire network.

\begin{figure}[!t]
\centering
\includegraphics[width=0.48\textwidth]{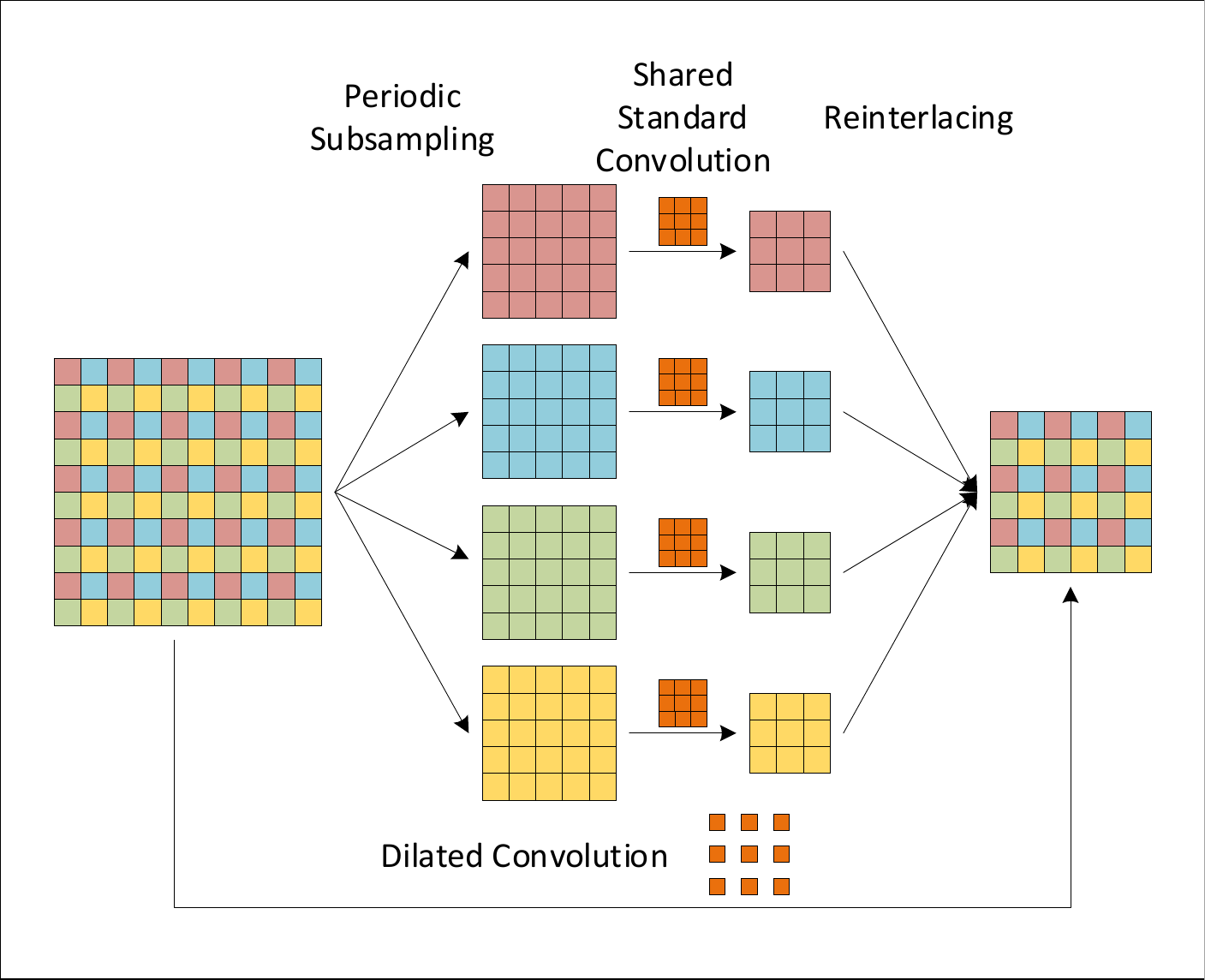}
\caption{An example of the decomposition of a dilated convolution
	with a kernel size of $3 \times 3$ and a dilation rate of $r=2$ on a
	$2$-D input feature map. The decomposition has three steps; namely
	periodic subsampling, shared standard convolution and reinterlacing.
	This example will also be used in Figures~\ref{fig:decompose}
	to~\ref{fig:method2b}.}
\label{fig:decompose}
\end{figure}

\subsection{A Decomposition View of Dilated Convolutions}\label{sec:decompose}

There are two ways to understand dilated convolutions. As introduced
in Section~\ref{sec:dilatedconv}, the first and more intuitive way
is to think of dilated convolutional filters with dilation rate $r$
as upsampled standard convolutional filters, by inserting
zeros~(holes)~\cite{papandreou2015modeling}. Another way to view
dilated convolutions is based on a decomposition of the
operation~\cite{shensa1992discrete}. A dilated convolution with a
dilation rate of $r$ can be decomposed into three steps. First, the
input feature maps are periodically subsampled by a factor of $r$.
As a result, the inputs are deinterlaced to $r^d$ groups of feature
maps of reduced resolution, where $d$ is the spatial dimension of
the inputs. Second, these groups of intermediate feature maps are
fed into a standard convolution. This convolution has filters with
the same weights as the original dilated convolution after removing
all inserted zeros. More importantly, it is shared for all the
groups, which means each group of reduced resolution maps goes through
the same standard convolution. The third step is to reinterlace the
$r^d$ groups of feature maps to the original resolution and produce
the outputs of the dilated convolution.

Figure~\ref{fig:decompose} gives an example of the decomposition in the
$2$-D case. To simplify the discussion, we assume the number of input
channels and output channels is both $1$. Given a $10 \times 10$
feature map, a dilated convolution with a kernel size of $3 \times
3$ and a dilation rate of $r=2$ will output a $6 \times 6$ feature
map without any padding. In the decomposition of this dilated
convolution, the input feature map is periodically subsampled into
$2^2=4$ groups of $5 \times 5$ feature maps of reduced resolution.
Then a shared standard convolution, which has the same weights as
the dilated convolution without padding, is applied to these $4$
groups of feature maps and obtains $4$ groups of $3 \times 3$
feature maps. Finally, they are reinterlaced to the original
resolution and produce exactly the same $6 \times 6$ output feature
map as the original dilated convolution. This decomposition reduces
dilated convolutions into standard convolutions and allows more
efficient
implementation~\cite{giusti2013fast,sermanet2013overfeat,chen2016deeplab,abadi2016tensorflow}.

We notice that the decomposition view provides a clear explanation
of the gridding artifacts; that is, the $r^d$ groups of intermediate
feature maps, either before or after the shared standard
convolution, have no dependency among each other and thus collect
potentially inconsistent local information. Based on this insight,
we overcome gridding by adding dependencies among the $r^d$ groups
in different steps of the decomposition. We propose two effective
approaches in the next two sections.

\begin{figure}[!t]
\centering
\includegraphics[width=0.48\textwidth]{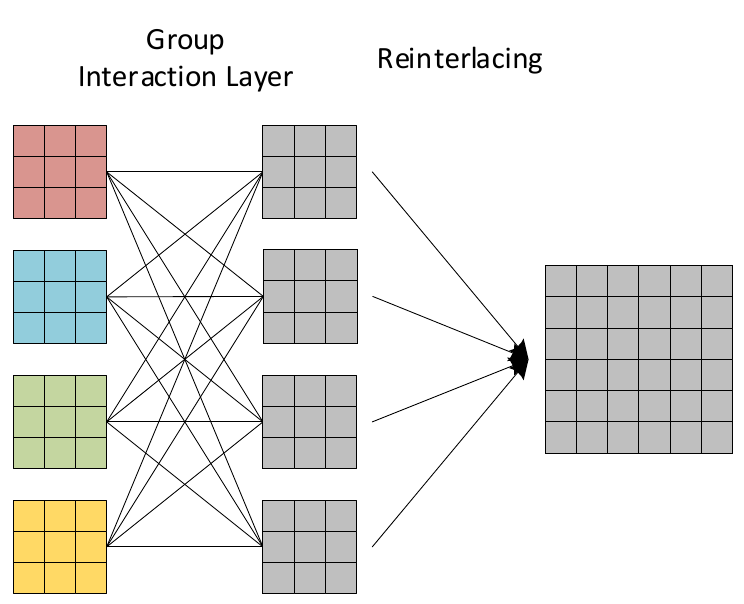}
\caption{Illustration of the degridding method in
	Section~\ref{sec:method1} for a dilated convolution with a kernel
	size of $3 \times 3$ and a dilation rate of $r=2$ on a $2$-D input
	feature map. By using a group interaction layer before
	reinterlacing, dependencies among intermediate groups are
	established. The same gray color denotes consistent local
	information.}
\label{fig:method1}
\end{figure}

\subsection{Smoothed Dilated Convolutions by Group Interaction Layers}\label{sec:method1}

Our first degridding method attempts to build dependencies among
different groups in the third step of the decomposition. We propose
to add a group interaction layer before reinterlacing the
intermediate feature maps to the original resolution. For a dilated
convolution with a dilation rate of $r$ on $d$-dimensional input
feature maps, the second step of the decomposition produces $r^d$
groups of feature maps of reduced resolution, denoted as
$\{f_i\}_{i=1}^{r^d}$, after the shared convolution. Note that each
$f_i$ represents a group of feature maps, rather than a single
feature map. We define a group interaction layer with a
weight matrix $W \in \mathbb{R}^{r^d \times r^d}$ given as
\begin{equation}\label{eqn:fcweights}
W=
\begin{bmatrix}
w_{11}    & w_{12}    & w_{13}    & \dots  & w_{1,r^d}  \\
w_{21}    & w_{22}    & w_{23}    & \dots  & w_{2,r^d}  \\
\vdots    & \vdots    & \vdots    & \ddots & \vdots     \\
w_{r^d,1} & w_{r^d,2} & w_{r^d,3} & \dots  & w_{r^d,r^d}
\end{bmatrix}.
\end{equation}
The outputs of this layer are still $r^d$ groups of feature maps,
denoted as $\{\hat{f}_i\}_{i=1}^{r^d}$, computed by
\begin{equation}\label{eqn:fcoutputs}
\hat{f}_i=\sum_{j=1}^{r^d}w_{ij} \cdot f_j,
\end{equation}
for $i=1,2,\ldots,r^d$.
Note that the connections of this layer are between groups instead of feature maps.
In fact, every $\hat{f}_i$ is a linear
combination of $\{f_i\}_{i=1}^{r^d}$, weighted by the weight matrix
$W$. Through this layer, each $\hat{f}_i$ collects local information
from all $r^d$ groups of feature maps, which adds dependencies among
different groups. After the group interaction layer, the
$r^d$ groups are reinterlaced to the original resolution and form
the final output of the dilated convolutions. The number of extra
training parameters in such smoothed dilated convolutions is
$r^{2d}$, independent of the number of input and output channels.
DCNNs with dilated convolutions are commonly used in one-dimensional
or two-dimensional cases, which means $d=1,2$. In practice, choices
of $r$ are usually $2,4,8$. The proposed group interaction
layer only requires learning thousands of extra parameters in the
worst cases, while the original dilated convolutions usually have
millions of training parameters.

We use the same example in Section~\ref{sec:decompose} to illustrate
the idea in Figure~\ref{fig:method1}. Given the outputs of the
second step in the decomposition, the $4$ groups of intermediate
feature maps build dependencies among each other through the
group interaction layer, whose number of weights is only
$2^{2 \cdot 2}=16$, represented by $16$ connections. We use the gray
color to represent feature maps after degridding.

\subsection{Smoothed Dilated Convolutions by Separable and Shared Convolutions}\label{sec:method2}

\begin{figure}[!t]
	\centering
	\includegraphics[width=0.48\textwidth]{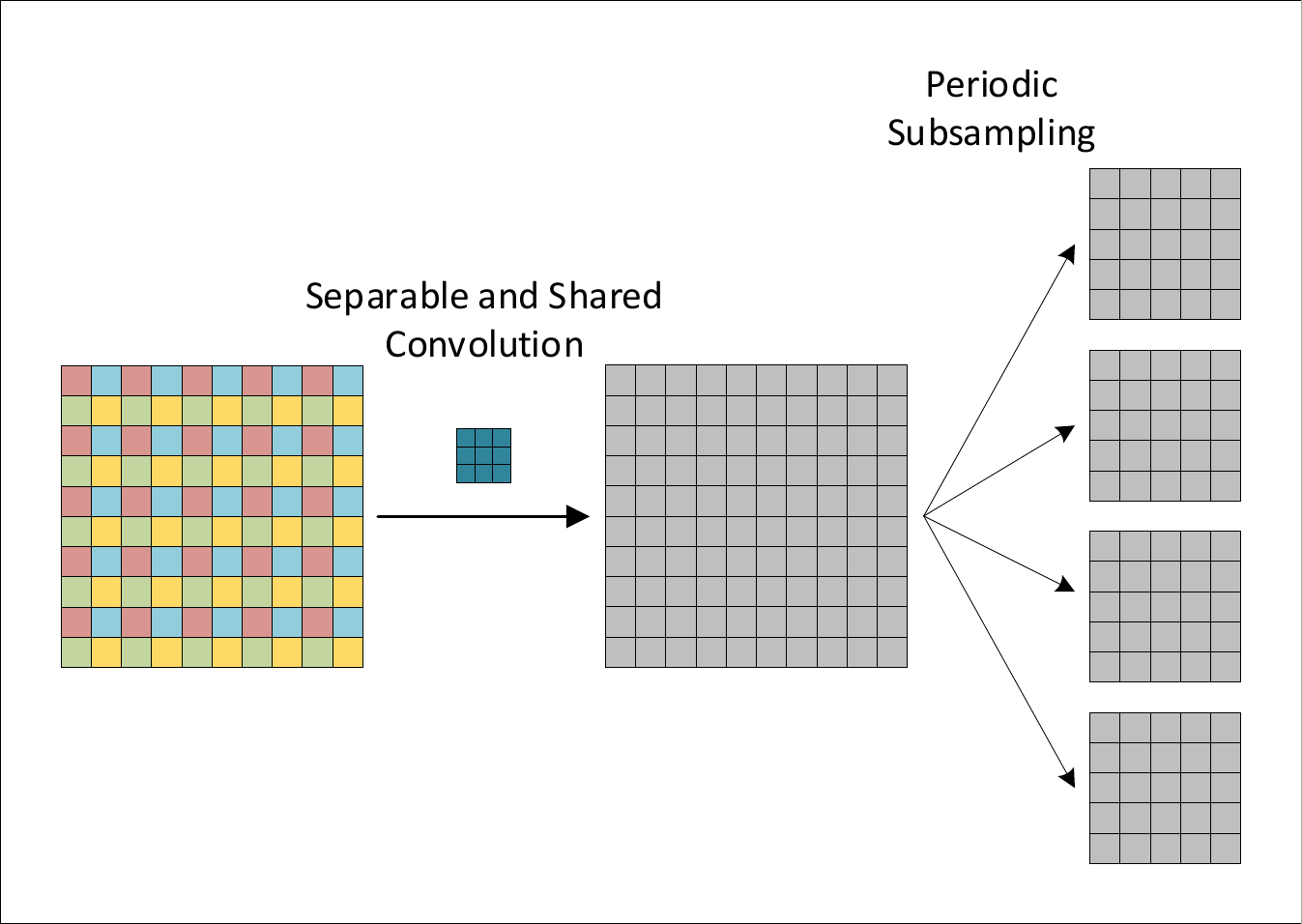}
	\caption{Illustration of the degridding method in
		Section~\ref{sec:method2} for a dilated convolution with a kernel
		size of $3 \times 3$ and a dilation rate of $r=2$ on a $2$-D input
		feature map. By adding the separable and shared convolution, the $4$
		groups created by periodic subsampling have dependencies among each
		other. The same gray color represents smoothed feature maps.}
	\label{fig:method2}
\end{figure}

\begin{figure*}[!t]
	\centering
	\includegraphics[width=0.8\textwidth]{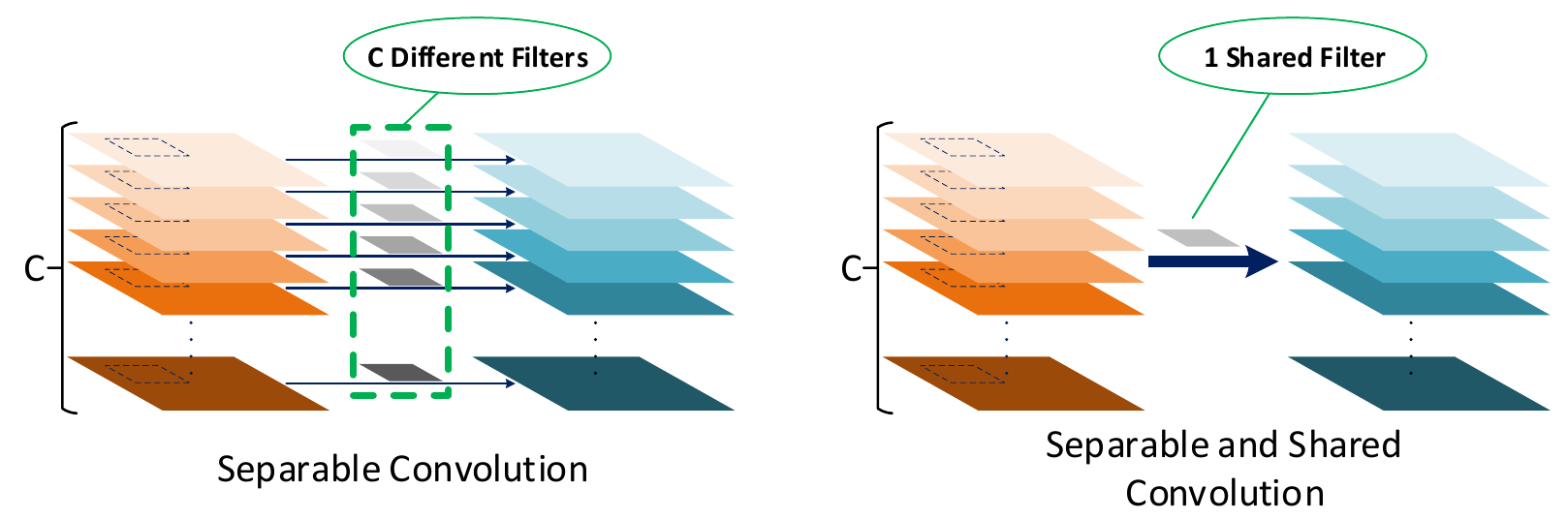}
	\caption{Illustration of the differences between the separable convolution and the proposed SS convolution introduced in Section~\ref{sec:method2}. For inputs and outputs of
		$C$ channels, the separable convolution has $C$ filters in total, with one filter for each channel, while the SS convolution only has one filter shared for all channels.}
	\label{fig:ssconv}
\end{figure*}

We further explore an approach to establish dependencies among
different groups in the first step of the decomposition; that is,
before deinterlacing the input feature maps. Considering a dilated
convolution with a dilation rate of $r$ on $d$-dimensional input
feature maps, the periodic subsampling during deinterlacing
distributes each unit in a local area of size $r^d$ in the inputs to
a separate group. Therefore, for units in a particular group, all
the neighboring units are in the other independent $r^d-1$ groups,
thereby resulting in local inconsistency. If the local information
can be incorporated before periodic sampling, it is possible to
alleviate the gridding artifacts.

In order to achieve this, we
propose separable and shared~(SS) convolutions, based on separable
convolutions~\cite{mamalet2012simplifying,chollet2016xception}.
Given inputs of $C$ channels and corresponding outputs of $C$
channels, separable convolutions are the same as standard
convolutions, except that separable convolutions handle each channel
separately. Standard convolutions connect all $C$ channels in inputs
to all $C$ channels in outputs, leading to $C^2$ different filters.
In contrast, separable convolutions only connect the $i$th output
channel to the $i$th input channel, yielding only $C$ filters. In
the proposed SS convolutions, ``shared'' means that, based on
separable convolutions, the $C$ filters are the same and shared by
all pairs of input and output channels. For inputs and outputs of
$C$ channels, SS convolutions only have one filter scanning all
spatial locations and share this filter across all channels.
Figure~\ref{fig:ssconv} provides a comparison between separable convolutions and SS convolutions.
In terms of smoothing dilated convolutions, we apply SS convolutions
to incorporate neighboring information for each unit in the input
feature maps. Specifically, an SS convolution with a kernel size of
$(2r-1)^d$ is inserted before deinterlacing, thereby adding
dependencies among each other to the $r^d$ groups of feature maps
produced by periodic subsampling.

The example in Figure~\ref{fig:method2} illustrates the idea of inserting SS
convolutions. Here, the kernel size of the inserted SS convolution is $(2
\cdot 2 -1)^2=3 \times 3$. Note that because the inputs only have one
channel, SS convolutions, separable convolutions and standard convolutions
are equivalent in this example. However, they become different if the inputs
have $C>1$ channels. Importantly, for inputs with multiple channels, the
number of training parameters does not change for SS convolutions, as opposed
to the other two kinds of convolutions. It means the proposed degridding
method has $(2r-1)^d$ parameters, independent of the number of channels,
which corresponds to only tens of extra parameters at most in practice.

\subsection{Relationship between the Two Methods}\label{sec:connection}

\begin{figure}[!t]
\centering
\includegraphics[width=0.48\textwidth]{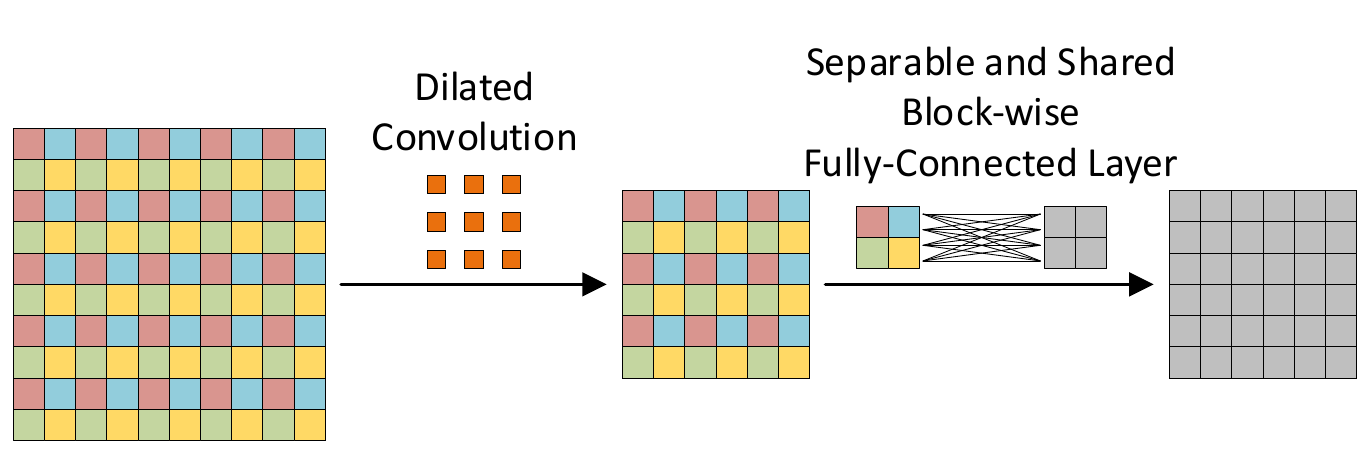}
\caption{Another illustration of the proposed method in
	Section~\ref{sec:method1}, corresponding to
	Figure~\ref{fig:method1}. The method is equivalent to adding an SS
	block-wise fully-connected layer after the dilated convolution.}
\label{fig:method1b}
\end{figure}

\begin{figure}[!t]
\centering
\includegraphics[width=0.48\textwidth]{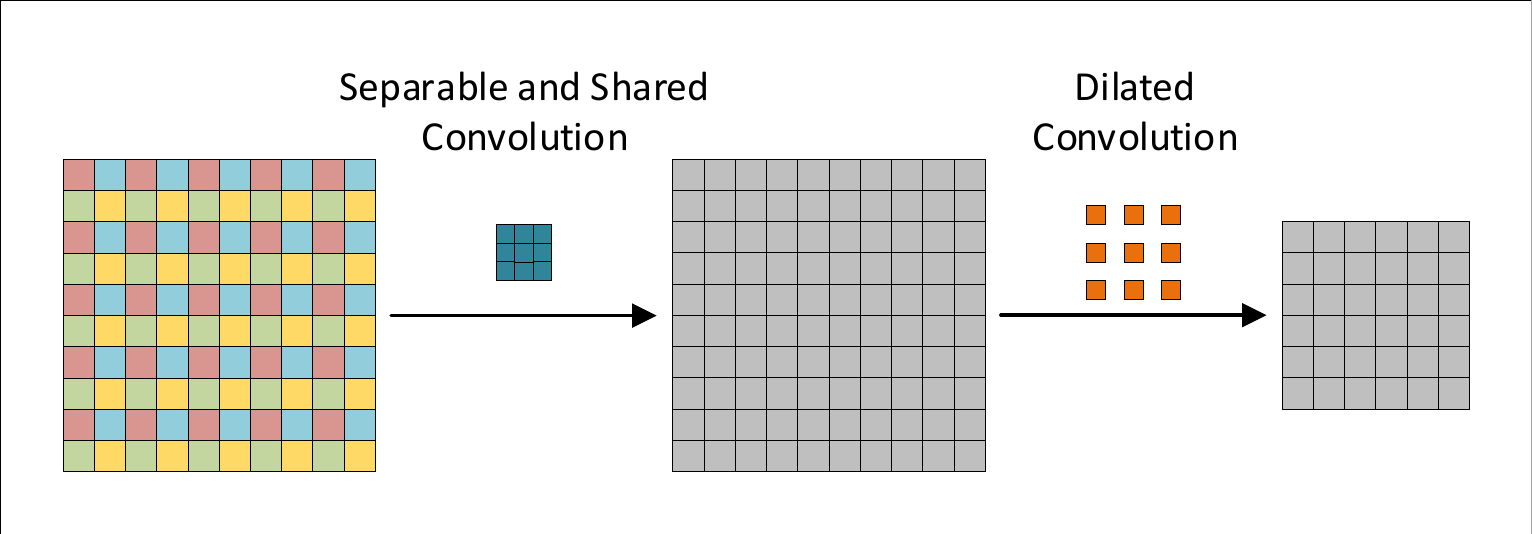}
\caption{Another illustration of the proposed method in
	Section~\ref{sec:method2}, corresponding to
	Figure~\ref{fig:method2}. The method is equivalent to inserting an
	SS convolution before the dilated convolution.}
\label{fig:method2b}
\end{figure}

\begin{figure*}[!t]
	\centering
	\includegraphics[width=0.95\textwidth]{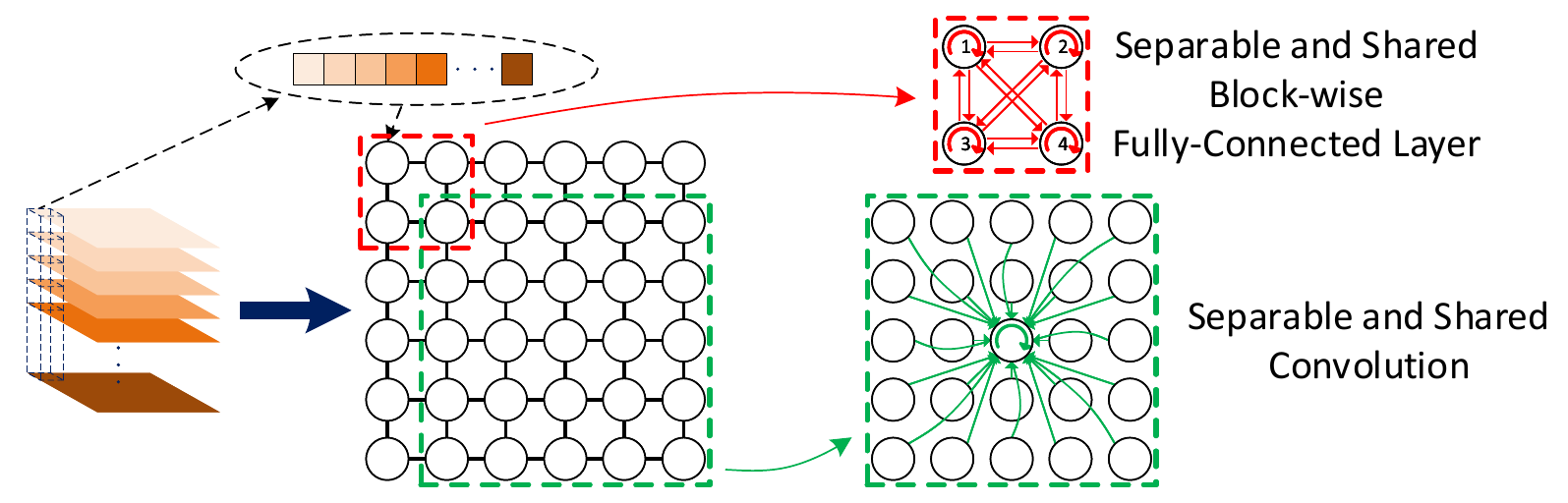}
	\caption{Illustration of SS operations in view of operations on graphs. Details are provided in Section~\ref{sec:ssop}. The circle arrow inside a node represents the self-loop.}
	\label{fig:transform}
\end{figure*}

Both of the proposed approaches are derived from the decomposition
view of dilated convolutions. Now we combine all steps and analyze
them in view of the original operation. For the second method in
Section~\ref{sec:method2}, it is straightforward as the separable
and shared~(SS) convolution is inserted before the first step of
decomposition and actually does not affect the original dilated
convolution. Consequently, it is equivalent to adding an SS
convolution before the dilated convolution, as shown in
Figure~\ref{fig:method2b}. However, the first method in
Section~\ref{sec:method1} performs degridding through the group-wise
fully-connected layer between the second and the third steps of the
decomposition. To see how to perform the combination, we refer to
the example in Figure~\ref{fig:method1}. Before the final step, we
have four groups of feature maps and each group has only one feature
map. Considering the units in the upper left corner of the four
feature maps, without the group interaction layer, these
four units form the upper left $2 \times 2$ block of the output
feature map after reinterlacing. If we insert the group-wise
fully-connected layer, the four new units in the upper left corner
become linear combinations of the previous ones and form the upper
left $2 \times 2$ block of the output feature map instead. As a
result, the new upper left $2 \times 2$ block of the output feature
map is computed by a fully-connected operation on the previous one.
By examining other units, we find that the fully-connected operation
is shared for every non-overlapping $2 \times 2$ blocks, scanning
the output feature map with a stride of $2$.
Figure~\ref{fig:method1b} provides an illustration. By generalizing
this example, we can see that the degridding method is equivalent to
a dilated convolution followed by the following operation: use a
window of size $r^d$ to scan the output feature map with stride $r$
and obtain non-overlapping blocks; for each block, perform the same
fully-connected operation that outputs a block of the same spatial
size. Note that if the outputs have multiple channels, the operation
is shared across channels. This operation is similar to the SS
convolution as they both scan spatial locations using a single
kernel shared across all channels. Thus, we name it as the SS
block-wise fully-connected layer. Based on it as well as the SS
convolution, we further define operations which scan spatial
locations of inputs using a single filter shared across all channels
as SS operations.

As DCNNs commonly employ dilated convolutional layers in cascade, we
also look into our proposed methods in this case. As explained
above, the first degridding approach is equivalent to adding an SS
block-wise fully-connected layer after the dilated convolution,
while the second one corresponds to inserting an SS convolution
before the dilated convolution. However, for a block of cascaded
dilated convolutional layers with the same dilation rate, the order
between the dilated convolution and the SS operation only affects
the very first and last layers. As a result, the two proposed
degridding methods can be generalized as combining appropriate SS
operations with dilated convolutions.

\subsection{Separable and Shared Operations}\label{sec:ssop}

\begin{figure*}[!t]
	\centering
	\includegraphics[width=0.9\textwidth]{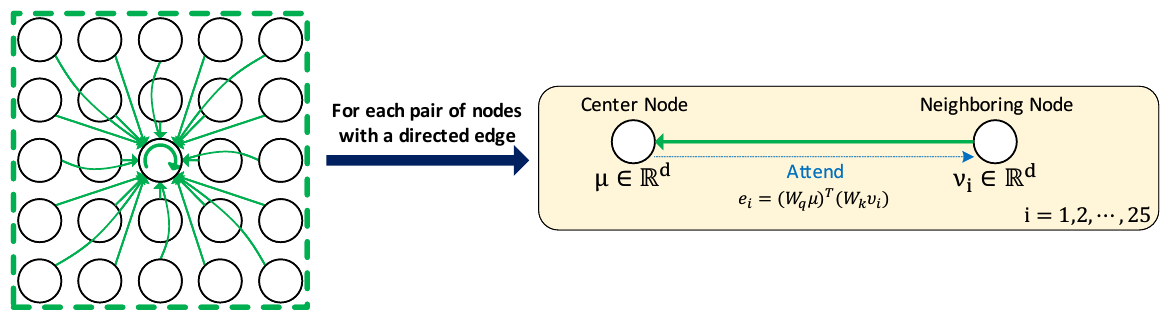}
	\caption{Illustration of our graph attention mechanism. Details are provided by Equations~\ref{eqn:att_coeff} to~\ref{eqn:att_sum} in Section~\ref{sec:smooth_output}. Here, $s=5$ so that $i=1,2,\ldots,25$.}
	\label{fig:graph_attend}
\end{figure*}

With the insights above, we develop more effective SS operations to
improve dense prediction models with dilated convolutions. According
to the definition in Section~\ref{sec:connection}, the key of SS
operations is to apply a filter that is shared across all channels.
Based on this property, we have reinvestigated SS operations in view
of operations on graphs. Note that the data that we focus on in this
work are grid-like data, such as 1-D text sequences, 2-D images, 3-D
videos, etc. For inputs of $C$ channels, each spatial location
corresponds to a $C$-dimensional vector. By treating each vector as
a node in a graph, the inputs are transformed into a grid-like
graph. The left part of Figure~\ref{fig:transform} provides an
illustration of this transformation for 2-D inputs. We first revisit
the proposed SS block-wise fully-connected layer and SS convolution
on this graph.


A $2 \times 2$ SS block-wise fully-connected layer scans the inputs
using a $2 \times 2$ window with a stride of $2$, as illustrated by
the red box in Figure~\ref{fig:transform}. To see the computation
within the window, we denote the four nodes as $n_1, n_2, n_3, n_4$
as marked in the figure. The filter $W \in \mathbb{R}^{4 \times 4}$
in this layer is given by
\begin{equation}
W=
\begin{bmatrix}
w_{11}    & w_{12}    & w_{13}   & w_{14} \\
w_{21}    & w_{22}    & w_{23}   & w_{24} \\
w_{31}    & w_{32}    & w_{33}   & w_{34} \\
w_{41}    & w_{42}    & w_{43}   & w_{44}
\end{bmatrix}.
\end{equation}
The outputs $\hat{n}_1, \hat{n}_2, \hat{n}_3, \hat{n}_4$ of the
window are computed by
\begin{equation}
\hat{n}_i = \sum_{j=1}^{4}w_{ij} \times n_j
\end{equation}
for $i=1,2,\ldots,4$. Here, $w_{ij} \times n_j$ means multiplying
each element of $n_j$ by $w_{ij}$, which is consistent with sharing
$W$ across all channels. In terms of operations on graphs, such
computation can be interpreted as a process on a directed subgraph
composed of the nodes in the scanning window. Every node interacts
with each other and produces its new representation, where the
interactions are modeled by directed edges. Specifically, the
subgraph does not follow the original grid-like connections.
Instead, each node has a directed edge to all nodes including
itself, as shown by the top right part of
Figure~\ref{fig:transform}. Each directed edge represents a scalar
weight in $W$. For example, the edge from $n_3$ to $n_1$ corresponds
to $w_{13}$, measuring the importance of $n_3$ to $\hat{n}_1$. In
other words, the SS block-wise fully-connected layer forms a
fully-connected directed subgraph in each window during scanning.

The SS convolution differs from the SS block-wise fully-connected
layer in that it constructs a different subgraph in its scanning
window. The bottom right part of Figure~\ref{fig:transform}
illustrates the directed subgraph for a $5 \times 5$ SS convolution.
Unlike the SS block-wise fully-connected layer, where all nodes in a
window get updated, the SS convolution only updates the
representation of the center node by incorporating information from
all nodes in the window. Therefore, the subgraph has directed edges
from all nodes to the center node. Nodes except for the center node
do not have self-loops or edges between each other. Again, each
directed edge refers to a scalar weight. There are 25 directed edges
corresponding to the $5 \times 5$ filter of the SS convolution.

To conclude, SS operations can be viewed as scanning the transformed
graph using a window. Within each window, a directed subgraph is
constructed, where each edge represents a scalar weight. Different
ways to form the directed graph result in different SS operations.
In addition, there are other ways to generate the scalar weights,
instead of making them as training parameters. Many studies on deep
learning on graphs have explored this direction, such as mixture
model networks~(MoNet)~\cite{monti2017geometric},
GraphSAGE~\cite{hamilton2017inductive}, graph attention
networks~(GAT)~\cite{velickovic2017graph}, and learnable graph
convolutional networks~(LGCN)~\cite{gao2018large}. In the next
section, we incorporate deep learning techniques on graphs and
propose an efficient and effective SS output layer, which improves
DCNNs with dilated convolutions by simply replacing the output
layer.

\subsection{Smoothed DCNNs with Dilated Convolutions}\label{sec:smooth_output}

The two proposed methods in Sections~\ref{sec:method1}
and~\ref{sec:method2} are able to smooth any single dilated
convolution. Our experimental results in
Section~\ref{sec:experiment} show that the proposed methods improve
the encoders of DCNNs with dilated convolutions. However, dilated
convolutions are also used in the output layer of these DCNNs. In
this section, we explore the use of SS operations to smooth the
entire network.

Various output layers have been proposed for DCNNs with dilated
convolutions, in order to aggregate information from large receptive
fields for prediction. For example, the large field of
view~(LargeFOV) layer in~\cite{chen2016deeplab} is a dilated
convolution with a kernel size of $3 \times 3$ and a dilation rate
of $r=12$ followed by $1 \times 1$ regular convolutions. The
LargeFOV layer has been extended to the atrous spatial pyramid
pooling~(ASPP) layer~\cite{chen2016deeplab,chen2017rethinking}. In
the ASPP layer, four LargeFOV layers with different dilation rates
are employed in parallel, and the outputs are summed or concatenated
together as the final output. However, both output layers do not
have any smoothing operation, thereby inheriting the gridding
artifacts from the encoder to the final output, as illustrated by
Figure~\ref{fig:gridding} in Section~\ref{sec:gridding}.

To address this problem, we propose the SS output layer, which
improves the performance by simply replacing dilated convolutions in
the output layer by an appropriate SS operation. The proposed SS
output layer is able to perform both smoothing and information
aggregation for prediction. First, in Section~\ref{sec:connection},
we conclude that the proposed degridding methods can be generalized
as inserting SS operations between consecutive dilated convolutions.
However, for DCNNs whose encoders use dilated convolutions in
cascade, it may be more efficient to add only one SS operation after
the entire encoder, making it a part of the output layer. Second,
the analysis in Section~\ref{sec:ssop} indicates that SS operations
are able to aggregate information within each scanning window. The
advantage of SS operations as compared with output layers based on
dilated convolutions is that, given the same receptive field,
information from all locations will be incorporated, instead of
sampled ones. In addition, SS operations usually have much fewer
parameters than dilated convolutions, as analyzed in
Section~\ref{sec:method2}. As a result, using the SS output layer is
efficient and effective.

To be specific, we first transform the output feature maps of the
encoder to the grid-like graph as shown in the left part of
Figure~\ref{fig:transform}. Then, we propose an SS operation that
constructs the same subgraph within each window as the SS
convolution, which is illustrated by the bottom right part of
Figure~\ref{fig:transform}. Differently, we adopt the graph
attention mechanism in GAT~\cite{velickovic2017graph} to generate
the scalar weights. Suppose the window size of our SS operation is
$s \times s$. There will be $s^2$ directed edges in the subgraph
constructed by the scanning window. We denote the starting nodes of
these directed edges as neighboring nodes $\upsilon_i,
i=1,2,\ldots,s^2$ and the center node as $\mu$. Note that
neighboring nodes include the center node. $\upsilon_i$ and $\mu$
are $d$-dimensional vectors, where $d$ is the number of input
channels. For each directed edge, we compute attention coefficients
defined as
\begin{equation}\label{eqn:att_coeff}
e_i=(W_q\mu)^T(W_k\upsilon_i),
\end{equation}
for $i=1,2,\ldots,s^2$. Here, $W_q, W_k \in \mathbb{R}^{d_k \times
	d}$ are shared for each edge and $d_k$ is a hyperparameter. The
attention coefficients are normalized across $i$, \emph{i.e.,} all
edges:
\begin{equation}\label{eqn:att_soft}
\alpha_i=Softmax(\frac{e_i}{\sqrt{d_k}}),
\end{equation}
where $\alpha_i$ is the generated scalar weight corresponding to the
$i$-th directed edge. The output of this window, which is the
updated representation of the center node, is computed by
\begin{equation}\label{eqn:att_sum}
\hat{\mu}=\sum_{i=1}^{s^2}\alpha_i \times W_v\upsilon_i,
\end{equation}
where $W_v \in \mathbb{R}^{d_o \times d}$ is also shared for each
edge and $d_o$ is a hyperparameter representing the dimension of
$\hat{\mu}$. Figure~\ref{fig:graph_attend} illustrates the graph
attention process within a $5 \times 5$ scanning window. Note that
if we choose the window size to be larger than the spatial sizes of
inputs, the SS operation is able to aggregate global information for
prediction. In addition, the SS operation has the same number of
parameters when changing the window size, because $W_q, W_k, W_v$
are all shared for edges, and different window sizes only result in
different number of edges.

In our SS output layer, the proposed SS operation scans the
transformed grid-like graph and updates every node. Appropriate
padding is employed and $\alpha_i$ is always set to $0$ for padding
nodes. We also apply the multi-head attention as in
GAT~\cite{velickovic2017graph}. A $1 \times1$ regular convolution
follows the SS operation to produce the final output. The proposed
SS output layer is evaluated in Section~\ref{sec:output}.

\section{Experimental Studies}\label{sec:experiment}

In this section, we evaluate our methods on the PASCAL
VOC~2012~\cite{everingham2010pascal} and
Cityscapes~\cite{cordts2016cityscapes} datasets. Our proposed
approaches result in significant and consistent improvements for
DCNNs with dilated convolutions. We also perform the effective
receptive field~(ERF) analysis~\cite{luo2016understanding} to
visualize the smoothing effect.
Finally, we analyze the effectiveness and efficiency of the proposed separable and shared output layer.

\subsection{Basic Setup}\label{sec:setup}

To conduct our experiments, we choose the task of semantic image
segmentation because the gridding artifacts were mainly observed in
studies for this
task~\cite{yu2017dilated,wang2017understanding,hamaguchi2017effective}.
The consistency of local information is important for such a
pixel-wise prediction task on images. In addition, the smoothing
effect is easy to visualize on two-dimensional data.

The baseline model in our experiments is the
DeepLabv2~\cite{chen2016deeplab} with ResNet-101~\cite{he2016deep}.
It is a fair benchmark to evaluate our smoothed dilated convolutions in three
aspects. First, it employed dilated convolutions to adapt ResNet
pre-trained on ImageNet~\cite{deng2009imagenet}; namely from image
classification to semantic image segmentation. Most semantic image
segmentation models adopted this transfer learning
strategy~\cite{giusti2013fast,li2014highly,long2015fully,yu2015multi,yu2017dilated,chen2016deeplab,chen2017rethinking,wang2017understanding,hamaguchi2017effective,zhao2016pyramid}
and ResNet is one of the most accurate DCNNs for image
classification with pre-trained models available. Second, models
that achieved the state-of-the-arts in segmentation tasks
recently~\cite{chen2017rethinking,wang2017understanding,zhao2016pyramid}
were developed from DeepLabv2. In~\cite{zhao2016pyramid} the output
layer was replaced with a pyramid pooling module.
\cite{wang2017understanding} also changed the output layer and
additionally proposed changing dilation rates, as mentioned in
Section~\ref{sec:gridding}. The current best model~\cite{chen2017rethinking} followed the
suggestions of~\cite{wang2017understanding} and meanwhile, explored
going deeper with more dilated convolutional blocks. Third, we
intend to compare our degridding methods with existing
approaches~\cite{yu2017dilated,hamaguchi2017effective,wang2017understanding}.
While~\cite{yu2017dilated,hamaguchi2017effective} addressed the
gridding artifacts by adding more layers that considerably increased
the number of training parameters, our methods only require learning
hundreds of extra parameters. Thus, we perform the comparison with
the idea proposed in~\cite{wang2017understanding}, which is based on
DeepLabv2.

DeepLabv2 is composed of two parts: the encoder and the output
layers. The encoder is a pre-trained ResNet-101 model modified with
dilated convolutions, and it extracts feature maps from raw images.
As introduced in Section~\ref{sec:dilatedconv}, the last two
downsampling layers in ResNet-101 were removed and subsequent
standard convolutional layers were replaced by dilated convolutional
layers with dilation rates of $r=2,4$, respectively. To be specific,
after the modification, the last two blocks are a block of $23$
stacked dilated convolutional layers with a dilation rate of $r=2$
followed by a block of $3$ cascaded dilated convolutions with a
dilation rate of $r=4$. The output layer performs pixel-wise
classification by aggregating information from the output feature
maps of encoder.

We re-implement DeepLabv2 in Tensorflow and perform experimental
studies based on our implementation. Our code is publicly
available\footnote{\url{https://github.com/divelab/dilated/}}. We
improve the baseline by addressing the gridding artifacts in the
last two blocks of the encoder. To make the comparison independent
of the output layer, we conduct experiments with different output
layers. In order to eliminate the bias of different datasets, we
evaluate our methods on two datasets. All the models are evaluated
by pixel intersection over union~(IoU) defined as
\begin{equation}\label{eqn:iou}
IoU = \frac{true\_positive}{true\_positive + false\_positive +
	false\_negative}.
\end{equation}

\subsection{PASCAL VOC2012}\label{sec:pascal}

\begin{table*}[h]
	\centering
	\caption{Experimental results of models with the ASPP output layer and
		MS-COCO pre-training on PASCAL VOC 2012 \textit{val} set. Class $1$ is the
		background class and Class $2-21$ represent ``aeroplane, bicycle, bird,
		boat, bottle, bus, car, cat, chair, cow, diningtable, dog, horse,
		motorbike, person, potteplant, sheep, sofa, train,
		tvmonitor'', respectively. This is the same for Tables~\ref{tab:resultp1}
		to~\ref{tab:resultp5}.}
	\label{tab:resultp1}
	\tabcolsep=0.1cm \begin{tabular}{l|ccccccccccccccccccccc|c}
		\toprule
		\textbf{Models} & \textbf{1} & \textbf{2} & \textbf{3} & \textbf{4} & \textbf{5} & \textbf{6} & \textbf{7} & \textbf{8} & \textbf{9} & \textbf{10} & \textbf{11} & \textbf{12} & \textbf{13} & \textbf{14} & \textbf{15} & \textbf{16} & \textbf{17} & \textbf{18} & \textbf{19} & \textbf{20} & \textbf{21} & \textbf{mIoU} \\
		\midrule
		DeepLabv2  & 93.8 & 85.9 & 38.8 & 84.8 & 64.3 & \textbf{79.0} & 93.7 & 85.5 & \textbf{91.7} & 34.1 & 83.0 & 57.0 & \textbf{86.1} & 83.0 & 81.0 & 85.0 & 58.2 & 83.4 & \textbf{48.2} & \textbf{87.2} & 74.0 & 75.1 \\
		\midrule
		Multigrid  & 93.6 & 85.4 & 38.9 & 82.2 & 66.9 & 76.6 & 93.2 & 85.3 & 90.7 & 35.7 & 82.5 & 53.7 & 83.1 & \textbf{84.2} & 82.2 & 84.6 & 56.9 & 84.3 & 45.6 & 85.5 & 73.1 & 74.5 \\
		\midrule
		G Interact & 93.7 & \textbf{86.9} & \textbf{39.6} & 84.1 & \textbf{68.9} & 76.4 & \textbf{93.8} & 86.2 & \textbf{91.7} & \textbf{36.1} & \textbf{83.7} & 55.3 & 85.7 & 84.0 & 82.2 & 84.9 & \textbf{59.5} & \textbf{85.7} & 46.5 & 85.0 & 73.0 & \textbf{75.4} \\
		SS Conv    & \textbf{93.9} & 86.7 & 39.5 & \textbf{86.2} & 68.1 & 77.3 & \textbf{93.8} & \textbf{86.4} & 91.5 & 35.4 & 83.2 & \textbf{59.0} & 85.2 & 83.6 & \textbf{82.4} & \textbf{85.2} & 57.3 & 82.1 & 45.8 & 86.1 & \textbf{75.2} & \textbf{75.4} \\
		\bottomrule
	\end{tabular}
\end{table*}

\begin{table*}[h]
	\centering
	\caption{Experimental results of models with the ASPP output layer but no
		MS-COCO pre-training on PASCAL VOC 2012 \textit{val} set.}
	\label{tab:resultp2}
	\tabcolsep=0.1cm \begin{tabular}{l|ccccccccccccccccccccc|c}
		\toprule
		\textbf{Models} & \textbf{1} & \textbf{2} & \textbf{3} & \textbf{4} & \textbf{5} & \textbf{6} & \textbf{7} & \textbf{8} & \textbf{9} & \textbf{10} & \textbf{11} & \textbf{12} & \textbf{13} & \textbf{14} & \textbf{15} & \textbf{16} & \textbf{17} & \textbf{18} & \textbf{19} & \textbf{20} & \textbf{21} & \textbf{mIoU} \\
		\midrule
		DeepLabv2  & 92.9 & 85.0 & 38.1 & 82.8 & 66.2 & 76.5 & 91.1 & 82.7 & 88.4 & \textbf{33.8} & 77.7 & 49.9 & 80.7 & 78.6 & 77.9 & 82.0 & 51.5 & 76.6 & 43.1 & 82.8 & 66.6 & 71.7 \\
		\midrule
		Multigrid  & 92.8 & 84.9 & 37.4 & 81.8 & 65.6 & 76.0 & 90.4 & 81.3 & 86.9 & 32.6 & 76.8 & 52.3 & 80.2 & 79.5 & 77.4 & 81.9 & 50.7 & 78.4 & 41.9 & 82.7 & 66.0 & 71.3 \\
		\midrule
		G Interact & \textbf{93.0} & 85.1 & 37.4 & \textbf{83.4} & \textbf{66.9} & 76.6 & 90.7 & 82.0 & 88.1 & \textbf{33.8} & \textbf{81.1} & \textbf{54.3} & 81.6 & \textbf{80.2} & 76.7 & 81.9 & \textbf{53.7} & \textbf{78.7} & 43.1 & \textbf{83.9} & 66.4 & \textbf{72.3} \\
		SS Conv    & \textbf{93.0} & \textbf{85.8} & \textbf{38.3} & 82.5 & 66.3 & \textbf{77.9} & \textbf{91.6} & \textbf{83.5} & \textbf{88.5} & 32.4 & 77.8 & 52.5 &	\textbf{81.9} & 78.1 & \textbf{79.3} & \textbf{82.1} & 49.8 & 78.4 & \textbf{44.4} & 83.0 & \textbf{67.9} & 72.1 \\
		\bottomrule
	\end{tabular}
\end{table*}

\begin{table*}[t]
	\centering
	\caption{Experimental results of models with the LargeFOV output layer and
		MS-COCO pre-training on PASCAL VOC 2012 \textit{val} set.}
	\label{tab:resultp3}
	\tabcolsep=0.1cm \begin{tabular}{l|ccccccccccccccccccccc|c}
		\toprule
		\textbf{Models} & \textbf{1} & \textbf{2} & \textbf{3} & \textbf{4} & \textbf{5} & \textbf{6} & \textbf{7} & \textbf{8} & \textbf{9} & \textbf{10} & \textbf{11} & \textbf{12} & \textbf{13} & \textbf{14} & \textbf{15} & \textbf{16} & \textbf{17} & \textbf{18} & \textbf{19} & \textbf{20} & \textbf{21} & \textbf{mIoU} \\
		\midrule
		DeepLabv2  & 93.7 & \textbf{85.7} & 39.4 & 85.9 & 67.6 & \textbf{79.0} & 93.1 & 86.0 & 90.7 & 36.2 & 79.8 & 54.6 & 83.7 & 80.9 & \textbf{81.4} & 85.0 & 57.5 & 83.5 & 45.5 & 84.5 & 74.1 & 74.7 \\
		\midrule
		G Interact & \textbf{93.8} & 85.5 & \textbf{40.0} & 86.5 & 67.5 & 78.1 & 92.9 & 86.2 & 90.4 & \textbf{37.2} & 80.6 & \textbf{56.5} & 82.6 & 80.3 & 81.0 & 85.0 & 58.1 & \textbf{84.8} & \textbf{46.6} & 84.4 & 74.8 & 74.9 \\
		SS Conv    & \textbf{93.8} & 85.3 & 39.7 & \textbf{86.8} & \textbf{68.7} & 77.9 & \textbf{94.0} & \textbf{86.3} & \textbf{90.8} & 35.2 & \textbf{83.1} & 55.4 & \textbf{84.5} & \textbf{83.8} & 79.6 & \textbf{85.6} & \textbf{59.3} & 83.2 & 46.2 & \textbf{86.2} & \textbf{75.5} & \textbf{75.3} \\
		\bottomrule
	\end{tabular}
\end{table*}

\begin{table*}[t]
	\centering
	\caption{Experimental results of models with the LargeFOV output layer but no
		MS-COCO pre-training on PASCAL VOC 2012 \textit{val} set.}
	\label{tab:resultp4}
	\tabcolsep=0.1cm \begin{tabular}{l|ccccccccccccccccccccc|c}
		\toprule
		\textbf{Models} & \textbf{1} & \textbf{2} & \textbf{3} & \textbf{4} & \textbf{5} & \textbf{6} & \textbf{7} & \textbf{8} & \textbf{9} & \textbf{10} & \textbf{11} & \textbf{12} & \textbf{13} & \textbf{14} & \textbf{15} & \textbf{16} & \textbf{17} & \textbf{18} & \textbf{19} & \textbf{20} & \textbf{21} & \textbf{mIoU} \\
		\midrule
		DeepLabv2  & 92.8 & 84.1 & 37.9 & 82.9 & 65.2 & \textbf{76.5} & 89.9 & 82.7 & 87.9 & 33.2 & 74.9 & 50.2 & 80.6 & 76.6 & 78.6 & 82.1 & 52.2 & 77.4 & 40.8 & 80.1 & 66.6 & 71.1 \\
		\midrule
		G Interact & \textbf{93.0} & 84.5 & 37.8 & \textbf{84.2} & \textbf{66.5} & 75.9 & 90.5 & 83.1 & \textbf{88.4} & \textbf{34.6} & \textbf{75.4} & \textbf{52.3} & \textbf{81.7} & 75.5 & 77.4 & 82.1 & 52.8 & 78.2 & 41.5 & \textbf{81.7} & \textbf{67.9} & \textbf{71.7} \\
		SS Conv    & 92.9 & \textbf{85.5} & \textbf{38.1} & 83.2 & \textbf{66.5} & 73.1 & \textbf{91.2} & \textbf{84.0} & 88.3 & 34.5 & 75.2 & 49.9 & 81.0 & \textbf{77.2} & \textbf{79.5} & \textbf{82.5} & \textbf{53.7} & \textbf{78.6} & \textbf{42.0} & 80.0 & 67.7 & 71.6 \\
		\bottomrule
	\end{tabular}
\end{table*}

The PASCAL VOC~2012 semantic image segmentation
dataset~\cite{everingham2010pascal} provides pixel-wise annotated
natural images. It has been split into \textit{train}, \textit{val}
and \textit{test} sets with $1,464$, $1,449$ and $1,456$ images,
respectively. The annotations include $21$ classes, which are $20$
foreground object classes and $1$ class for background. An augmented
version with extra annotations~\cite{hariharan2011semantic}
increases the size of the \textit{train} set to $10,582$. In our
experiments, we train all the models using the augmented
\textit{train} set and evaluate them on the \textit{val} set. When
reproducing the baseline DeepLabv2, we do not employ multi-scale
inputs with max fusion for testing due to our limited GPU memory. We
perform no post-processing such as conditional random
fields~(CRF)~\cite{chen2016deeplab}, which is not related to our
goals. Following DeepLabv2, we train the model with randomly cropped
patches of size of $321 \times 321$ and batch size of $10$. Data
augmentation by randomly scaling the inputs for training is applied.
We set the initial learning rate to $0.00025$ and adopt the ``poly''
learning rate policy~\cite{liu2015parsenet} as
\begin{equation}\label{eqn:poly}
current\_lr = (1-\frac{iter}{max\_iter})^{power} \cdot initial\_lr,
\end{equation}
where $power=0.9$, $iter$ denotes current iteration number, and $lr$
denotes learning rate, as
in~\cite{chen2016deeplab,chen2017rethinking,wang2017understanding}.
The model is trained for $max\_iter=20,000$ iterations with a
momentum of $0.9$ and a weight decay of $0.0005$.

We implement our proposed methods by inserting appropriate separable
and shared~(SS) operations before or after each dilated convolution
as shown in Figures~\ref{fig:method1b} and~\ref{fig:method2b}. An
important step is to change the initial learning rate, detailed in
each experiment. To make the comparisons solid, we also train the
baseline with different initial learning rates and observe the
original setting of $0.00025$ yields the best performance. The
initialization of SS operations is to set them to be identity
operations. Specifically, for a group interaction layer
with a dilation rate of $r=2$, the initial filter is
\begin{equation}\label{eqn:fcinit}
W=
\begin{bmatrix}
1    & 0    & 0    & 0 \\
0    & 1    & 0    & 0 \\
0    & 0    & 1    & 0 \\
0    & 0    & 0    & 1
\end{bmatrix},
\end{equation}
while for an SS convolution with a dilation rate of $r=2$, it is
\begin{equation}\label{eqn:convinit}
W=
\begin{bmatrix}
0    & 0    & 0 \\
0    & 1    & 0 \\
0    & 0    & 0
\end{bmatrix}.
\end{equation}

The original DeepLabv2 used pre-training on
MS-COCO~\cite{lin2014microsoft}, which results in more training data
and higher performances. Our experiments are conducted under both
settings; namely with and without MS-COCO pre-training. The results
are given in Tables~\ref{tab:resultp1} and~\ref{tab:resultp2},
respectively. In the tables, ``G Interact'' denotes the degridding method
with a group interaction layer, {\emph i.e.,} adding an SS block-wise fully-connected layer after the dilated convolution
and ``SS Conv'' represents the one with an SS
convolution inserted before the dilated convolution. In these
experiments with MS-COCO pre-training, the initial learning rates
for ``G Interact'' and ``SS Conv'' are both $0.001$. Otherwise, they are
set to $0.001$ and $0.00075$, respectively. Clearly, both proposed
methods improve the IoU for most classes as well as the mean
IoU~(mIoU) over the baseline under both settings. It is worth noting
that ``G Interact'' only requires training $1,136(=16 \times 23 + 256
\times 3)$ extra parameters and ``SS Conv'' requires $354(=9 \times
23 + 49 \times 3)$ extra parameters, which are negligible compared
to the total number of parameters in the models.

\begin{table*}[t]
\centering
\caption{Experimental results of models with the ASPP output layer and
	MS-COCO pre-training on Cityscapes \textit{val} set. Class $1-19$ represent
	``road, sidewalk, building, wall, fence, pole, traffic light, traffic sign,
	vegetation, terrain, sky, person, rider, car, truck, bus, train,
	motorcycle, bicycle'', respectively. This is the same for
	Tables~\ref{tab:resultc1} and~\ref{tab:resultc2}.}
\label{tab:resultc1}
\tabcolsep=0.1cm \begin{tabular}{l|ccccccccccccccccccc|c}
\toprule
\textbf{Models} & \textbf{1} & \textbf{2} & \textbf{3} & \textbf{4} & \textbf{5} & \textbf{6} & \textbf{7} & \textbf{8} & \textbf{9} & \textbf{10} & \textbf{11} & \textbf{12} & \textbf{13} & \textbf{14} & \textbf{15} & \textbf{16} & \textbf{17} & \textbf{18} & \textbf{19} & \textbf{mIoU} \\
\midrule
DeepLabv2  & 97.2 & \textbf{79.7} & 90.1 & 47.4 & 49.2 & 50.3 & 57.3 & 69.0 & \textbf{90.6} & 59.8 & \textbf{92.8} & 75.9 & 55.6 & 92.5 & 67.5 & 80.5 & 64.8 & 59.7 & 71.7 & 71.1 \\
\midrule
G Interact & \textbf{97.3} & 79.6 & 90.2 & 50.4 & 49.9 & \textbf{50.5} & \textbf{58.5} & 69.1 & 90.5 & 58.7 & 92.7 & 75.9 & 55.4 & 92.5 & 70.9 & 80.2 & 65.0 & \textbf{60.6} & \textbf{71.8} & 71.6 \\
SS Conv    & 97.2 & \textbf{79.7} & \textbf{90.3} & \textbf{51.1} & \textbf{50.5} & 50.2 & 58.1 & \textbf{69.3} & 90.5 & \textbf{60.0} & 92.7 & \textbf{76.1} & \textbf{55.9} & \textbf{92.7} & \textbf{72.7} & \textbf{81.9} & \textbf{66.0} & 59.7 & \textbf{71.8} & \textbf{71.9} \\
\bottomrule
\end{tabular}
\end{table*}

\begin{table*}[t]
\centering
\caption{Experimental results of models with the ASPP output layer but no
	MS-COCO pre-training on Cityscapes \textit{val} set.}
\label{tab:resultc2}
\tabcolsep=0.1cm \begin{tabular}{l|ccccccccccccccccccc|c}
\toprule
\textbf{Models} & \textbf{1} & \textbf{2} & \textbf{3} & \textbf{4} & \textbf{5} & \textbf{6} & \textbf{7} & \textbf{8} & \textbf{9} & \textbf{10} & \textbf{11} & \textbf{12} & \textbf{13} & \textbf{14} & \textbf{15} & \textbf{16} & \textbf{17} & \textbf{18} & \textbf{19} & \textbf{mIoU} \\
\midrule
DeepLabv2  & 97.0 & 77.9 & 89.4 & 44.6 & 48.6 & 48.7 & 54.1 & 66.7 & \textbf{90.3} & \textbf{58.0} & \textbf{92.5} & 73.9 & 51.9 & 91.6 & \textbf{59.9} & 75.5 & 60.5 & 56.3 & 69.6 & 68.8 \\
\midrule
G Interact & \textbf{97.1} & \textbf{78.7} & \textbf{89.6} & 44.7 & 49.2 & 48.6 & 54.2 & \textbf{67.0} & \textbf{90.3} & 57.6 & 92.1 & \textbf{74.3} & \textbf{52.2} & 91.7 & 59.0 & \textbf{77.1} & 60.5 & 56.8 & \textbf{70.1} & 69.0 \\
SS Conv    & 97.0 & 78.3 & \textbf{89.6} & \textbf{45.2} & \textbf{49.4} & \textbf{48.9} & \textbf{54.6} & 66.5 & 90.2 & 57.1 & 92.0 & 74.1 & 52.1 & \textbf{91.8} & 59.5 & 76.8 & \textbf{63.5} & \textbf{58.8} & 69.7 & \textbf{69.2} \\
\bottomrule
\end{tabular}
\end{table*}

We also compare our methods with existing degridding method proposed
in~\cite{wang2017understanding} and used
in~\cite{chen2017rethinking} as the ``multigrid'' method. As
introduced in Section~\ref{sec:gridding}, the idea is to group
several dilated convolutional layers and change the dilation
factors. As we know, for the modified ResNet-101 with dilated
convolutions, the last two blocks are a block of $23$ stacked
dilated convolutional layers with a dilation rate of $r=2$ followed
by a block of $3$ cascaded dilated convolutions with a dilation rate
of $r=4$. For the first block, we group every $3$ layers together
and replace the dilation rates from $r=2,2,2$ to $r=1,2,3$. We keep
$r=2,2$ for the left $2$ layers. For the second block, the $3$
dilation factors $r=4,4,4$ are changed to $r=3,4,5$. We make the
modification and train the models under the same setting as the
baseline. The results, denoted as ``Multigrid'', are shown in the
second lines of Tables~\ref{tab:resultp1} and~\ref{tab:resultp2}.
Surprisingly, our implementation indicates that the approach does
not improve the performance. An explanation of the results is that
the method should be applied together with other modifications, as
both~\cite{wang2017understanding} and~\cite{chen2017rethinking}
conduct experiments together with other changes over DeepLabv2, such
as dense upsamling convolution~(DUC) and deeper encoders.

As we address the gridding artifacts in the last two blocks of the
encoder, we also run experiments with different output layers in
order to make the comparisons independent of the output layer. We
replace the original atrous spatial pyramid pooling~(ASPP) output
layer of DeepLabv2 by the large field of view~(LargeFOV) layer,
which was applied earlier in~\cite{chen2016deeplab}. We train the
models with the same settings above, with and without MS-COCO
pre-training, and show the results in Tables~\ref{tab:resultp3}
and~\ref{tab:resultp4}, respectively. Again, the proposed degridding
methods result in significant improvements consistently.

\subsection{Cityscapes}\label{sec:city}

We further compare our proposed methods on the Cityscapes
dataset~\cite{cordts2016cityscapes}. Cityscapes collects $5,000$
$2048 \times 1024$ images of street scenes from $50$ different
cities and provides high quality pixel-wise annotations of $19$
classes. The $5,000$ images are divided into \textit{train},
\textit{val} and \textit{test} with $2,975$, $500$ and $1,525$
images, respectively. Again, we train models on the \textit{train}
set and perform evaluation on the \textit{val} set. The training
batch size is $3$, where each batch contains randomly cropped
patches of size $571 \times 571$. The initial learning rates for all
models are set to $0.0005$. All the other settings are the same as
those in Section~\ref{sec:pascal}.

Experiments are still conducted under both settings, {\emph i.e.,}
with and without MS-COCO pre-training, and the results are given in
Tables~\ref{tab:resultc1} and~\ref{tab:resultc2}, respectively. We
can see that both of the proposed methods increase the mIoU over the
baseline, which shows that the improvements are independent of
datasets.

\subsection{Effective Receptive Field Analysis}\label{sec:ERF}

Since we are addressing the gridding artifacts, we perform the
effective receptive field~(ERF)
analysis~\cite{luo2016understanding,hamaguchi2017effective} to
visualize the smoothing effect of our methods. These experiments
further verify that the improvements of the proposed methods come
from degridding. Given a block in DCNNs, the ERF analysis is an
approach to characterize how much each unit in the input of the
block affects a particular output unit of the block
mathematically~\cite{luo2016understanding}, instead of
theoretically.

\begin{figure}[!t]
	\centering
	\includegraphics[width=0.48\textwidth]{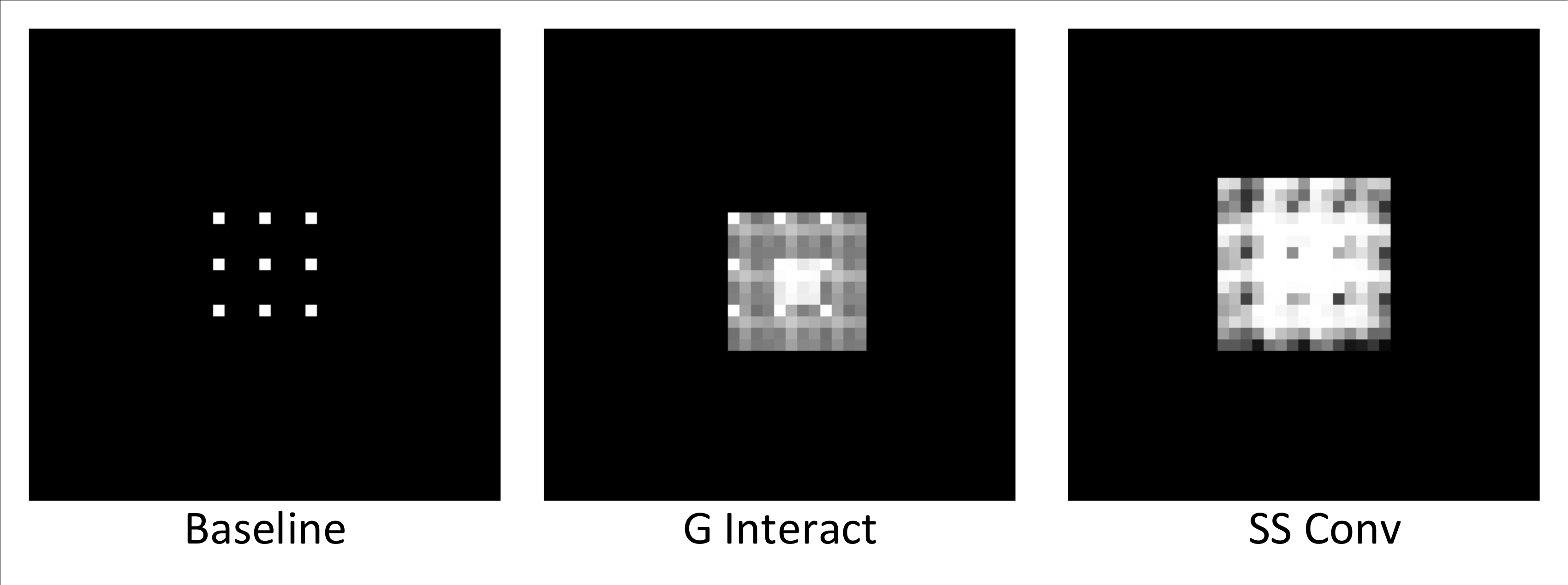}
	\caption{ERF visualization for the single dilated convolution with a
		kernel size of $3 \times 3$ and a dilation factor of $r=4$. Black
		pixels represent zero weights.}
	\label{fig:resulterf1}
\end{figure}

\begin{figure}[!t]
	\centering
	\includegraphics[width=0.48\textwidth]{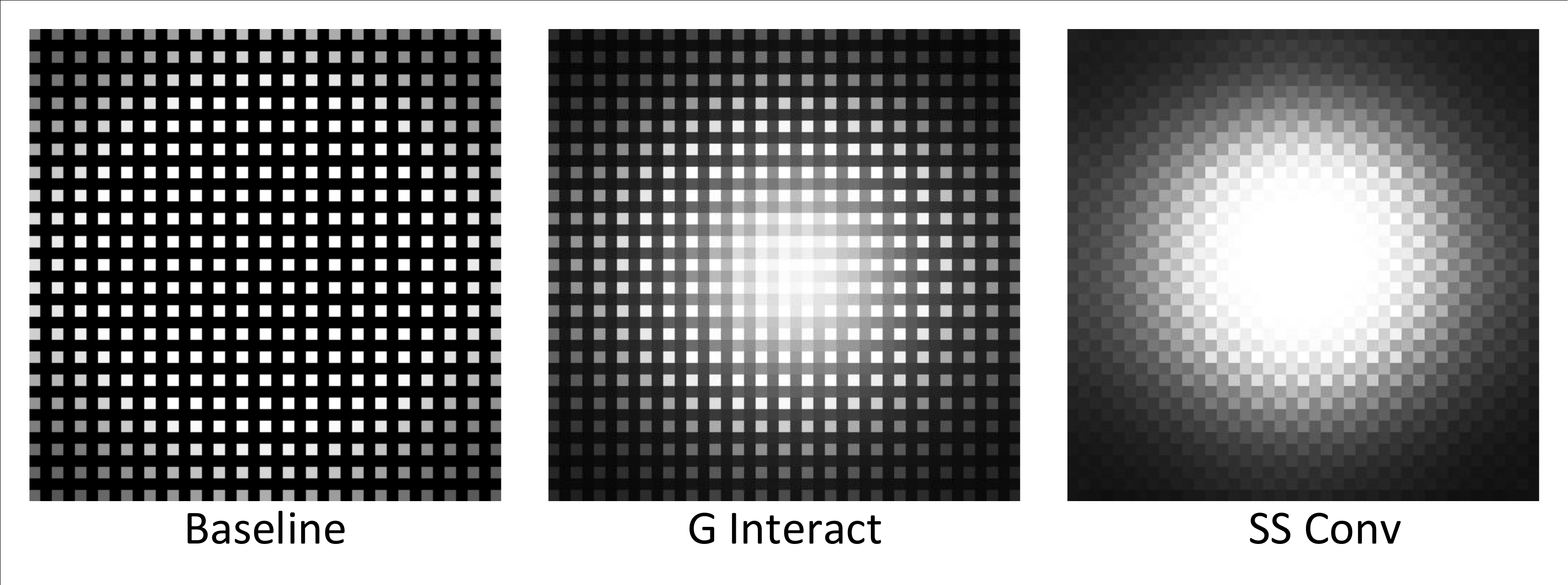}
	\caption{ERF visualization for the entire dilated convolutional
		block. Note that only the leftmost map has black pixels that
		represent zero weights.}
	\label{fig:resulterf2}
\end{figure}

\begin{table*}[h]
	\centering
	\caption{Experimental results of models with the proposed SS output layer in Section~\ref{sec:smooth_output} and
		MS-COCO pre-training on PASCAL VOC 2012 \textit{val} set. ``SS Output ($\#$)'' denotes the model using an SS output layer with a window size of $\# \times \#$. ``SS Output Global'' means the window size is chosen to be larger than the spatial sizes of inputs.}
	\label{tab:resultp5}
	\tabcolsep=0.085cm \begin{tabular}{l|ccccccccccccccccccccc|c}
		\toprule
		\textbf{Models} & \textbf{1} & \textbf{2} & \textbf{3} & \textbf{4} & \textbf{5} & \textbf{6} & \textbf{7} & \textbf{8} & \textbf{9} & \textbf{10} & \textbf{11} & \textbf{12} & \textbf{13} & \textbf{14} & \textbf{15} & \textbf{16} & \textbf{17} & \textbf{18} & \textbf{19} & \textbf{20} & \textbf{21} & \textbf{mIoU} \\
		\midrule
		DeepLabv2 (LargeFOV) & 93.7 & 85.7 & 39.4 & 85.9 & 67.6 & 79.0 & 93.1 & 86.0 & 90.7 & 36.2 & 79.8 & 54.6 & 83.7 & 80.9 & 81.4 & 85.0 & 57.5 & 83.5 & 45.5 & 84.5 & 74.1 & 74.7 \\
		DeepLabv2 (ASPP) & 93.8 & 85.9 & 38.8 & 84.8 & 64.3 & 79.0 & 93.7 & 85.5 & 91.7 & 34.1 & 83.0 & 57.0 & 86.1 & 83.0 & 81.0 & 85.0 & 58.2 & 83.4 & 48.2 & 87.2 & 74.0 & 75.1 \\
		\midrule
		SS Output (15)  & 94.1 & 86.5 & 39.0 & 86.2 & 65.9 & 80.3 & 93.8 & 87.4 & 90.7 & 36.0 & 82.1 & 59.2 & 84.2 & 80.8 & 81.2 & 85.5 & 58.3 & 84.1 & 48.1 & 87.4 & 74.1 & 75.5 \\
		SS Output (20)  & 94.1 & 86.7 & 40.2 & 86.9 & 66.2 & \textbf{80.5} & 94.5 & \textbf{87.9} & 91.7 & 36.1 & 83.6 & \textbf{60.0} & 86.2 & 83.9 & 82.5 & 85.6 & 58.9 & 84.0 & 48.9 & 88.7 & 74.3 & 76.3 \\
		SS Output (30)  & 94.2 & 87.9 & \textbf{40.9} & \textbf{87.3} & 66.1 & 79.7 & \textbf{94.8} & \textbf{87.9} & 92.9 & 36.5 & 84.9 & 59.6 & \textbf{88.3} & 86.8 & \textbf{83.9} & \textbf{85.7} & \textbf{59.9} & 84.2 & 50.0 & 89.5 & \textbf{74.9} & 77.0 \\
		\midrule
		SS Output Global & \textbf{94.4} & \textbf{89.2} & 40.6 & 84.9 & \textbf{69.7} & 78.9 & 94.7 & 86.8 & \textbf{93.2} & \textbf{38.1} & \textbf{89.9} & 59.4 & 87.8 & \textbf{87.3} & 82.6 & \textbf{85.7} & 59.4 & \textbf{89.5} & \textbf{52.4} & \textbf{90.2} & 74.7 & \textbf{77.6} \\
		\bottomrule
	\end{tabular}
\end{table*}

\begin{table}[h]
	\centering
	\caption{Comparison of the number of training parameters between different output layers. The numbers of channels in inputs and outputs are set to $2,048$ and $512$, respectively. Note that the SS output layer has the same number of parameters when changing the window size.}
	\label{tab:result_param}
	\begin{tabular}{lr}
		\toprule
		\textbf{Models} & \#Parameters \\
		\midrule
		LargeFOV & 9,437,696 \\
		ASPP & 37,750,784 \\
		SS Output & 3,409,920 \\
		\bottomrule
	\end{tabular}
\end{table}

Following the steps
in~\cite{luo2016understanding,hamaguchi2017effective}, we analyze
the models on PASCAL VOC 2012, with the ASPP output layer and
MS-COCO pre-training. We compute the ERF for chosen blocks of the
baseline and both of the proposed methods. Specifically, suppose the
input and output feature maps of a block are $x$ and $y$,
respectively. The spatial locations of the feature maps are indexed
by $(i,j)$ with $(0,0)$ representing the center. The ERF is measured
by the partial derivative $\partial y_{0,0} / \partial x_{i,j}$. To
compute it without an explicit loss function, we set the error
gradient with respect to $y_{0,0}$ to $1$ while for $y_{i,j}$ with
$i \neq 0$ or $j \neq 0$, we set it to $0$. Then the error gradient
can be back-propagated to $x$ and the error gradient with respect to
$x_{i,j}$ equals to $\partial y_{0,0} / \partial
x_{i,j}$~\cite{luo2016understanding}. However, the results are
input-dependent. So $\partial y_{0,0} / \partial x_{i,j}$ are
computed for all images in the \textit{val} set and their absolute
values are averaged. Finally, we sum the values over all channels of
$x$ to get a visualization of the ERF.

In our experiments, we choose two blocks of the DCNNs to visualize
the smoothing effect and enlarge the spatial size of visualizations
ten times for display. The first block is the very last layer of the
encoder, which is a dilated convolution with a kernel size of $3
\times 3$ and a dilation rate of $r=4$. The ERF analysis results are
presented in Figure~\ref{fig:resulterf1}. The ERF of the original
dilated convolution in the baseline is obvious. It corresponds to a
$3 \times 3$ filter with zeros inserted between non-zero weights.
Such a filter results in the gridding problem. For our proposed
degridding methods, we can see that they smooth the ERF and thus
perform degridding. In addition, both methods expand the rectangular
size of the ERF due to the SS operations. The second chosen block is
the entire block composed of dilated convolutional layers, which
includes the last two blocks of the encoder.
Figure~\ref{fig:resulterf2} shows the ERF visualization. The
gridding artifacts are clearly smoothed in both proposed methods. In
fact, only the leftmost visualization for the baseline has black
pixels that represent zero weights. Particularly, we note that ``SS
FC'' still has a grid-like visualization. A reason of this is the
block-wise operation may result in larger grids in terms of blocks.
Nevertheless, it alleviates the inconsistency of pixel-wise local
information and improves DCNNs with dilated convolutions.

\subsection{Separable and Shared Output Layer}\label{sec:output}

We evaluate the proposed separable and shared~(SS) output layer in
Section~\ref{sec:smooth_output} by only replacing the output layer
of DeepLabv2. In our experiments, we set $d_k$ and $d_o$ to $512$ in
Equations~\ref{eqn:att_coeff} to~\ref{eqn:att_sum} and use $8$ heads
in the graph attention mechanism. Different window sizes of the SS
output layer are explored. Table~\ref{tab:resultp5} provides the
comparison results between the original DeepLabv2 and models using
SS output layers. Clearly, the SS output layer shows its
effectiveness by improving the performance significantly. It is
worth noting that the larger the window size is, the more the
performance gets improved, which indicates the importance of
aggregating global information for prediction.

In order to show the efficiency of our SS output layer, we also
compare the number of training parameters between different output
layers in Table~\ref{tab:result_param}. The number of input channels
to the output layer is set to $2,048$. To be fair, we only compute
the number of parameters of the operations before the $1 \times 1$
regular convolutions and set the number of output channels of the
operations to $512$. In this case, the LargeFov output layer has a
single $3 \times 3$ dilated convolution and the ASPP output layer
has four $3 \times 3$ dilated convolutions, while the SS output
layer contains the proposed SS operation in
Section~\ref{sec:smooth_output} with $d_k=d_o=512$. Note that the SS
output layer has the same number of parameters when changing the
window size. According to Table~\ref{tab:result_param}, the proposed
SS output layer reduces a large amount of training parameters as
compared with output layers based on dilated convolutions.

\section{Conclusions}\label{sec:conclusion}

In this work, we propose two simple yet effective degridding methods
based on a decomposition of dilated convolutions. The proposed
methods differ from existing degridding approaches in two aspects.
First, we address the gridding artifacts in terms of a single
dilated convolution operation instead of multiple layers in cascade.
Second, our methods only require learning a negligible amount of
extra parameters. Experimental results show that they improve DCNNs
with dilated convolutions significantly and consistently. The
smoothing effect is also visualized in the effective receptive
field~(ERF) analysis. Through further analysis, we relate both
proposed methods together and define the separable and shared~(SS)
operations. The newly defined SS operation is a general neural
network operation and may result in a general degridding strategy.
We explore this direction in this updated version and propose the SS
output layer, which is able to smooth the entire network by only
replacing the output layer and obtain improved dense prediction. To
conclude, our proposed degridding methods based on SS operations are
efficient and effective.

\ifCLASSOPTIONcompsoc
  \section*{Acknowledgments}
\else
  \section*{Acknowledgment}
\fi

This work was supported in part by National Science Foundation grant
IIS-1633359 and Defense Advanced Research Projects Agency grant
N66001-17-2-4031.

\ifCLASSOPTIONcaptionsoff
  \newpage
\fi



\bibliographystyle{IEEEtran}
\bibliography{seg-reference}

\begin{thebibliography}{10}
\providecommand{\url}[1]{#1}
\csname url@samestyle\endcsname
\providecommand{\newblock}{\relax}
\providecommand{\bibinfo}[2]{#2}
\providecommand{\BIBentrySTDinterwordspacing}{\spaceskip=0pt\relax}
\providecommand{\BIBentryALTinterwordstretchfactor}{4}
\providecommand{\BIBentryALTinterwordspacing}{\spaceskip=\fontdimen2\font plus
\BIBentryALTinterwordstretchfactor\fontdimen3\font minus
  \fontdimen4\font\relax}
\providecommand{\BIBforeignlanguage}[2]{{%
\expandafter\ifx\csname l@#1\endcsname\relax
\typeout{** WARNING: IEEEtran.bst: No hyphenation pattern has been}%
\typeout{** loaded for the language `#1'. Using the pattern for}%
\typeout{** the default language instead.}%
\else
\language=\csname l@#1\endcsname
\fi
#2}}
\providecommand{\BIBdecl}{\relax}
\BIBdecl

\bibitem{giusti2013fast}
A.~Giusti, D.~C. Ciresan, J.~Masci, L.~M. Gambardella, and J.~Schmidhuber,
  ``Fast image scanning with deep max-pooling convolutional neural networks,''
  in \emph{Image Processing (ICIP), 2013 20th IEEE International Conference
  on}.\hskip 1em plus 0.5em minus 0.4em\relax IEEE, 2013, pp. 4034--4038.

\bibitem{li2014highly}
H.~Li, R.~Zhao, and X.~Wang, ``Highly efficient forward and backward
  propagation of convolutional neural networks for pixelwise classification,''
  \emph{arXiv preprint arXiv:1412.4526}, 2014.

\bibitem{yu2015multi}
F.~Yu and V.~Koltun, ``Multi-scale context aggregation by dilated
  convolutions,'' \emph{arXiv preprint arXiv:1511.07122}, 2015.

\bibitem{yu2017dilated}
F.~Yu, V.~Koltun, and T.~Funkhouser, ``Dilated residual networks,'' \emph{arXiv
  preprint arXiv:1705.09914}, 2017.

\bibitem{chen2016deeplab}
L.-C. Chen, G.~Papandreou, I.~Kokkinos, K.~Murphy, and A.~L. Yuille, ``Deeplab:
  Semantic image segmentation with deep convolutional nets, atrous convolution,
  and fully connected crfs,'' \emph{arXiv preprint arXiv:1606.00915}, 2016.

\bibitem{chen2017rethinking}
L.-C. Chen, G.~Papandreou, F.~Schroff, and H.~Adam, ``Rethinking atrous
  convolution for semantic image segmentation,'' \emph{arXiv preprint
  arXiv:1706.05587}, 2017.

\bibitem{wang2017understanding}
P.~Wang, P.~Chen, Y.~Yuan, D.~Liu, Z.~Huang, X.~Hou, and G.~Cottrell,
  ``Understanding convolution for semantic segmentation,'' \emph{arXiv preprint
  arXiv:1702.08502}, 2017.

\bibitem{hamaguchi2017effective}
R.~Hamaguchi, A.~Fujita, K.~Nemoto, T.~Imaizumi, and S.~Hikosaka, ``Effective
  use of dilated convolutions for segmenting small object instances in remote
  sensing imagery,'' \emph{arXiv preprint arXiv:1709.00179}, 2017.

\bibitem{zhao2016pyramid}
H.~Zhao, J.~Shi, X.~Qi, X.~Wang, and J.~Jia, ``Pyramid scene parsing network,''
  \emph{arXiv preprint arXiv:1612.01105}, 2016.

\bibitem{gao2017pixel}
H.~Gao, H.~Yuan, Z.~Wang, and S.~Ji, ``Pixel deconvolutional networks,''
  \emph{arXiv preprint arXiv:1705.06820}, 2017.

\bibitem{sermanet2013overfeat}
P.~Sermanet, D.~Eigen, X.~Zhang, M.~Mathieu, R.~Fergus, and Y.~LeCun,
  ``Overfeat: Integrated recognition, localization and detection using
  convolutional networks,'' \emph{arXiv preprint arXiv:1312.6229}, 2013.

\bibitem{papandreou2015modeling}
G.~Papandreou, I.~Kokkinos, and P.-A. Savalle, ``Modeling local and global
  deformations in deep learning: Epitomic convolution, multiple instance
  learning, and sliding window detection,'' in \emph{Proceedings of the IEEE
  Conference on Computer Vision and Pattern Recognition}, 2015, pp. 390--399.

\bibitem{dai2016r}
J.~Dai, Y.~Li, K.~He, and J.~Sun, ``R-fcn: Object detection via region-based
  fully convolutional networks,'' in \emph{Advances in neural information
  processing systems}, 2016, pp. 379--387.

\bibitem{huang2016speed}
J.~Huang, V.~Rathod, C.~Sun, M.~Zhu, A.~Korattikara, A.~Fathi, I.~Fischer,
  Z.~Wojna, Y.~Song, S.~Guadarrama \emph{et~al.}, ``Speed/accuracy trade-offs
  for modern convolutional object detectors,'' \emph{arXiv preprint
  arXiv:1611.10012}, 2016.

\bibitem{oord2016wavenet}
A.~v.~d. Oord, S.~Dieleman, H.~Zen, K.~Simonyan, O.~Vinyals, A.~Graves,
  N.~Kalchbrenner, A.~Senior, and K.~Kavukcuoglu, ``Wavenet: A generative model
  for raw audio,'' \emph{arXiv preprint arXiv:1609.03499}, 2016.

\bibitem{kalchbrenner2016video}
N.~Kalchbrenner, A.~v.~d. Oord, K.~Simonyan, I.~Danihelka, O.~Vinyals,
  A.~Graves, and K.~Kavukcuoglu, ``Video pixel networks,'' \emph{arXiv preprint
  arXiv:1610.00527}, 2016.

\bibitem{kalchbrenner2016neural}
N.~Kalchbrenner, L.~Espeholt, K.~Simonyan, A.~v.~d. Oord, A.~Graves, and
  K.~Kavukcuoglu, ``Neural machine translation in linear time,'' \emph{arXiv
  preprint arXiv:1610.10099}, 2016.

\bibitem{holschneider1990real}
M.~Holschneider, R.~Kronland-Martinet, J.~Morlet, and P.~Tchamitchian, ``A
  real-time algorithm for signal analysis with the help of the wavelet
  transform,'' in \emph{Wavelets}.\hskip 1em plus 0.5em minus 0.4em\relax
  Springer, 1990, pp. 286--297.

\bibitem{deng2009imagenet}
J.~Deng, W.~Dong, R.~Socher, L.-J. Li, K.~Li, and L.~Fei-Fei, ``Imagenet: A
  large-scale hierarchical image database,'' in \emph{Computer Vision and
  Pattern Recognition, 2009. CVPR 2009. IEEE Conference on}.\hskip 1em plus
  0.5em minus 0.4em\relax IEEE, 2009, pp. 248--255.

\bibitem{he2016deep}
K.~He, X.~Zhang, S.~Ren, and J.~Sun, ``Deep residual learning for image
  recognition,'' in \emph{Proceedings of the IEEE conference on computer vision
  and pattern recognition}, 2016, pp. 770--778.

\bibitem{long2015fully}
J.~Long, E.~Shelhamer, and T.~Darrell, ``Fully convolutional networks for
  semantic segmentation,'' in \emph{Proceedings of the IEEE Conference on
  Computer Vision and Pattern Recognition}, 2015, pp. 3431--3440.

\bibitem{shensa1992discrete}
M.~J. Shensa, ``The discrete wavelet transform: wedding the a trous and mallat
  algorithms,'' \emph{IEEE Transactions on signal processing}, vol.~40, no.~10,
  pp. 2464--2482, 1992.

\bibitem{abadi2016tensorflow}
M.~Abadi, A.~Agarwal, P.~Barham, E.~Brevdo, Z.~Chen, C.~Citro, G.~S. Corrado,
  A.~Davis, J.~Dean, M.~Devin \emph{et~al.}, ``Tensorflow: Large-scale machine
  learning on heterogeneous distributed systems,'' \emph{arXiv preprint
  arXiv:1603.04467}, 2016.

\bibitem{luo2016understanding}
W.~Luo, Y.~Li, R.~Urtasun, and R.~Zemel, ``Understanding the effective
  receptive field in deep convolutional neural networks,'' in \emph{Advances in
  Neural Information Processing Systems}, 2016, pp. 4898--4906.

\bibitem{wang2018smoothed}
Z.~Wang and S.~Ji, ``Smoothed dilated convolutions for improved dense
  prediction,'' in \emph{Proceedings of the 24th ACM SIGKDD International
  Conference on Knowledge Discovery \& Data Mining}.\hskip 1em plus 0.5em minus
  0.4em\relax ACM, 2018, pp. 2486--2495.

\bibitem{mamalet2012simplifying}
F.~Mamalet and C.~Garcia, ``Simplifying convnets for fast learning,''
  \emph{Artificial Neural Networks and Machine Learning--ICANN 2012}, pp.
  58--65, 2012.

\bibitem{chollet2016xception}
F.~Chollet, ``Xception: Deep learning with depthwise separable convolutions,''
  \emph{arXiv preprint arXiv:1610.02357}, 2016.

\bibitem{monti2017geometric}
F.~Monti, D.~Boscaini, J.~Masci, E.~Rodola, J.~Svoboda, and M.~M. Bronstein,
  ``Geometric deep learning on graphs and manifolds using mixture model cnns,''
  in \emph{Proc. CVPR}, vol.~1, no.~2, 2017, p.~3.

\bibitem{hamilton2017inductive}
W.~Hamilton, Z.~Ying, and J.~Leskovec, ``Inductive representation learning on
  large graphs,'' in \emph{Advances in Neural Information Processing Systems},
  2017, pp. 1024--1034.

\bibitem{velickovic2017graph}
P.~Velickovic, G.~Cucurull, A.~Casanova, A.~Romero, P.~Lio, and Y.~Bengio,
  ``Graph attention networks,'' \emph{arXiv preprint arXiv:1710.10903}, 2017.

\bibitem{gao2018large}
H.~Gao, Z.~Wang, and S.~Ji, ``Large-scale learnable graph convolutional
  networks,'' in \emph{Proceedings of the 24th ACM SIGKDD International
  Conference on Knowledge Discovery \& Data Mining}.\hskip 1em plus 0.5em minus
  0.4em\relax ACM, 2018, pp. 1416--1424.

\bibitem{everingham2010pascal}
M.~Everingham, L.~Van~Gool, C.~K. Williams, J.~Winn, and A.~Zisserman, ``The
  pascal visual object classes (voc) challenge,'' \emph{International journal
  of computer vision}, vol.~88, no.~2, pp. 303--338, 2010.

\bibitem{cordts2016cityscapes}
M.~Cordts, M.~Omran, S.~Ramos, T.~Rehfeld, M.~Enzweiler, R.~Benenson,
  U.~Franke, S.~Roth, and B.~Schiele, ``The cityscapes dataset for semantic
  urban scene understanding,'' in \emph{Proceedings of the IEEE Conference on
  Computer Vision and Pattern Recognition}, 2016, pp. 3213--3223.

\bibitem{hariharan2011semantic}
B.~Hariharan, P.~Arbel{\'a}ez, L.~Bourdev, S.~Maji, and J.~Malik, ``Semantic
  contours from inverse detectors,'' in \emph{Computer Vision (ICCV), 2011 IEEE
  International Conference on}.\hskip 1em plus 0.5em minus 0.4em\relax IEEE,
  2011, pp. 991--998.

\bibitem{liu2015parsenet}
W.~Liu, A.~Rabinovich, and A.~C. Berg, ``Parsenet: Looking wider to see
  better,'' \emph{arXiv preprint arXiv:1506.04579}, 2015.

\bibitem{lin2014microsoft}
T.-Y. Lin, M.~Maire, S.~Belongie, J.~Hays, P.~Perona, D.~Ramanan,
  P.~Doll{\'a}r, and C.~L. Zitnick, ``Microsoft coco: Common objects in
  context,'' in \emph{European conference on computer vision}.\hskip 1em plus
  0.5em minus 0.4em\relax Springer, 2014, pp. 740--755.

\end{thebibliography}
\end{document}